# SOCIALLY AWARE NAVIGATION FOR MOBILE ROBOTS: A SURVEY ON DEEP REINFORCEMENT LEARNING APPROACHES

**Ibrahim Khalil Kabir[1,2], Muhammad Faizan Mysorewala[1,2,3]**

[1] Control and Instrumentation Engineering Department, King Fahd University of Petroleum & Minerals, Dhahran 31261, Saudi Arabia

[2] Interdisciplinary Research Centre for Smart Mobility and Logistics, King Fahd University of Petroleum & Minerals, Dhahran 31261, Saudi Arabia

[3] Interdisciplinary Research Centre for Intelligent Manufacturing and Robotics, King Fahd University of Petroleum & Minerals, Dhahran 31261, Saudi Arabia

g202214320@kfupm.edu.sa, mysorewala@kfupm.edu.sa

## ABSTRACT

Socially aware navigation is a fast-evolving research area in robotics that enables robots to move within human environments while adhering to the implicit human social norms. The advent of Deep Reinforcement Learning (DRL) has accelerated the development of navigation policies that enable robots to incorporate these social conventions while effectively reaching their objectives. This survey offers a comprehensive overview of DRL-based approaches to socially aware navigation, highlighting key aspects such as proxemics, human comfort, naturalness, trajectory and intention prediction, which enhance robot interaction in human environments. This work critically analyzes the integration of value-based, policy-based, and actor-critic reinforcement learning algorithms alongside neural network architectures, such as feedforward, recurrent, convolutional, graph, and transformer networks, for enhancing agent learning and representation in socially aware navigation. Furthermore, we examine crucial evaluation mechanisms, including metrics, benchmark datasets, simulation environments, and the persistent challenges of sim-to-real transfer. Our comparative analysis of the literature reveals that while DRL significantly improves safety, and human acceptance over traditional approaches, the field still faces setback due to non-uniform evaluation mechanisms, absence of standardized social metrics, computational burdens that limit scalability, and difficulty in transferring simulation to real robotic hardware applications. We assert that future progress will depend on hybrid approaches that leverage the strengths of multiple approaches and producing benchmarks that balance technical efficiency with human-centered evaluation. By reviewing the state of the art and highlighting these challenges, this work provides researchers and practitioners with a roadmap for advancing DRL-based socially aware navigation toward robust, real-world deployment.

## 1. INTRODUCTION

The field of robotics has seen much development in the past decades, moving from simple robots to complex devices with decision-making capabilities and the ability to work in complex and dynamic environments [1]. Since robotics has seen such rapid development, it has moved from being used in a few environments to being applied in virtually all fields and areas of life. Robots are increasingly used in human environments, and interactions between them thus become necessary to define. This has led to the rise of a new field of research under robotics, commonly called social robot navigation, human-aware navigation, or socially aware navigation (SAN). Much research has been done in this field, especially in recent years, due to advancements in how robots are being used in public and human spaces, and because of the drive towards efficiency and improvement in the quality of life.

Socially aware navigation enables robots to navigate in dynamic and crowded environments by proactively planning interactions with people and navigation with human-like behaviour to ensure the comfort and safety of humans in the vicinity, as captured in the illustration in Figure 1. Socially aware navigation fundamentally incorporates proxemics and social norms into the navigation policies to avoid causing discomfort, fear, or danger to humans while trying to achieve its goals.

Socially aware navigation is one of a few terms used to refer to this field. Generally, they all mean the same thing unless explicitly stated by the author. The most commonly used are 'socially aware navigation', 'human-aware navigation', 'crowd-aware navigation', and 'social navigation'. 'Socially aware navigation' focuses on making robots adhere to social norms while moving within human environments. 'Human aware navigation' focuses on understanding, reacting, and manoeuvring around a human or a small group of humans while considering human social norms. 'Crowd-aware navigation' involves navigating socially through large groups of humans, usually densely situated, while also understanding their collective behaviour. 'Social navigation' is a broader term that may include all the above and involves all aspects of robot navigation within social contexts or norms. These social norms or mannerisms are typically subtle rules or behaviours of humans that govern human interaction, such as maintaining appropriate distances (proxemics), avoiding passing through humans engaged in a conversation, predicting human movement, and reacting to human gestures and cues. Some of these rules and considerations include the following, as proposed by [2]: safety, comfort, legibility, politeness, social competency, agent understanding, proactivity, and contextual appropriateness. This is demonstrated in Figure 1.

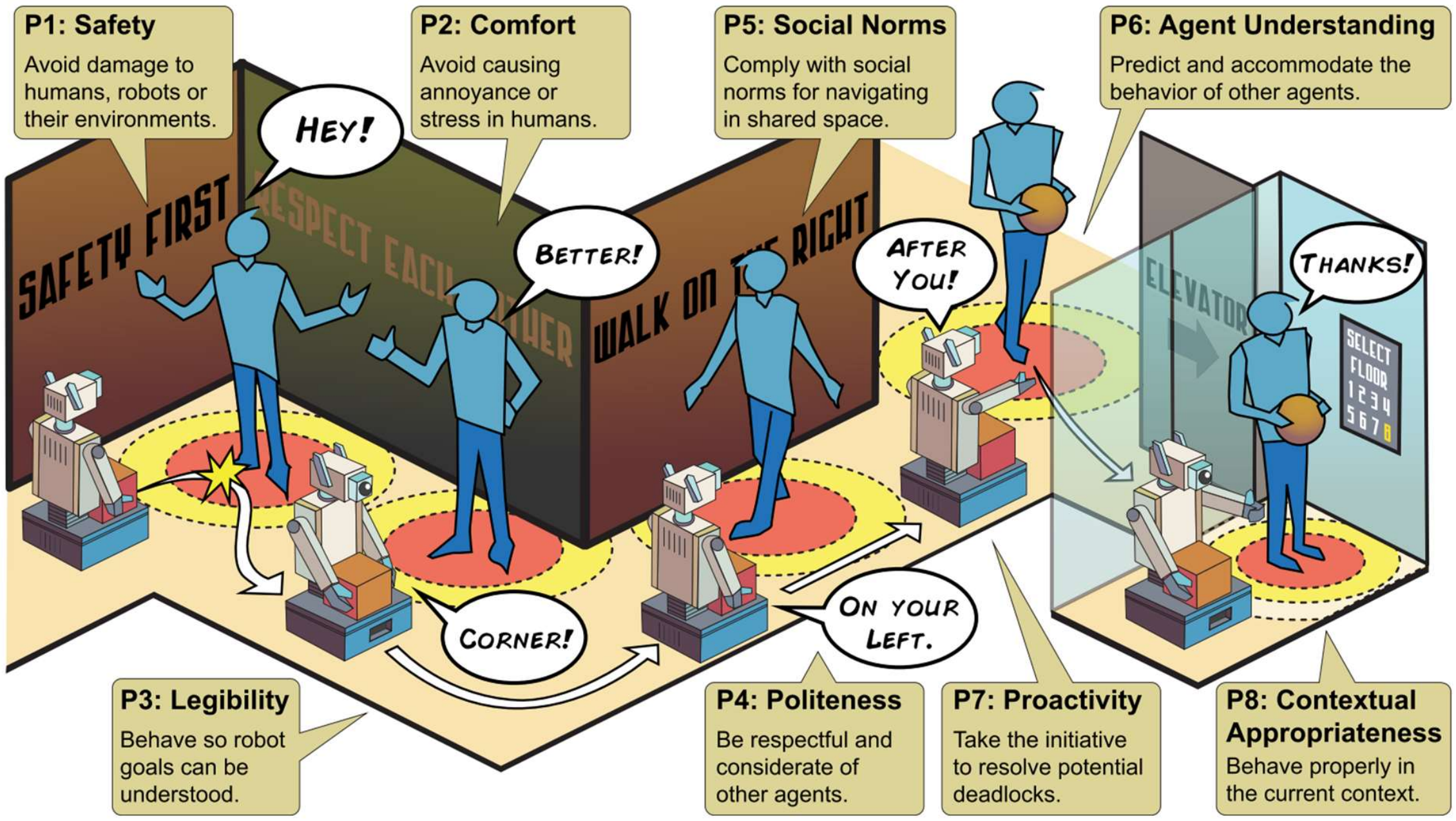

**Fig 1** Principles for social robot navigation [2].

Deep reinforcement learning (DRL) is a type of machine learning that combines reinforcement learning with Deep Neural Networks (DNNs) [3] to solve problems. DRL's ultimate goal is to learn a policy that maximises cumulative rewards over time using a process of trial and error coupled with feedback from the environment [4]. It uses deep learning techniques to handle high-dimensional state and action spaces, which may be difficult for traditional RL algorithms. This makes it popular in applications like robotics and renewable power systems [5].

DRL has found good applicability in socially aware navigation because it enhances the navigation of mobile robots by addressing the dynamics of complex environments and multiple navigation scenarios while considering social norms and human preferences. Using DRL has the advantage over traditional methods in that it can learn policies that consider not only the navigation goal but also proxemics, motion patterns, attention, and social conventions, and take into account interactions among humans themselves [6].

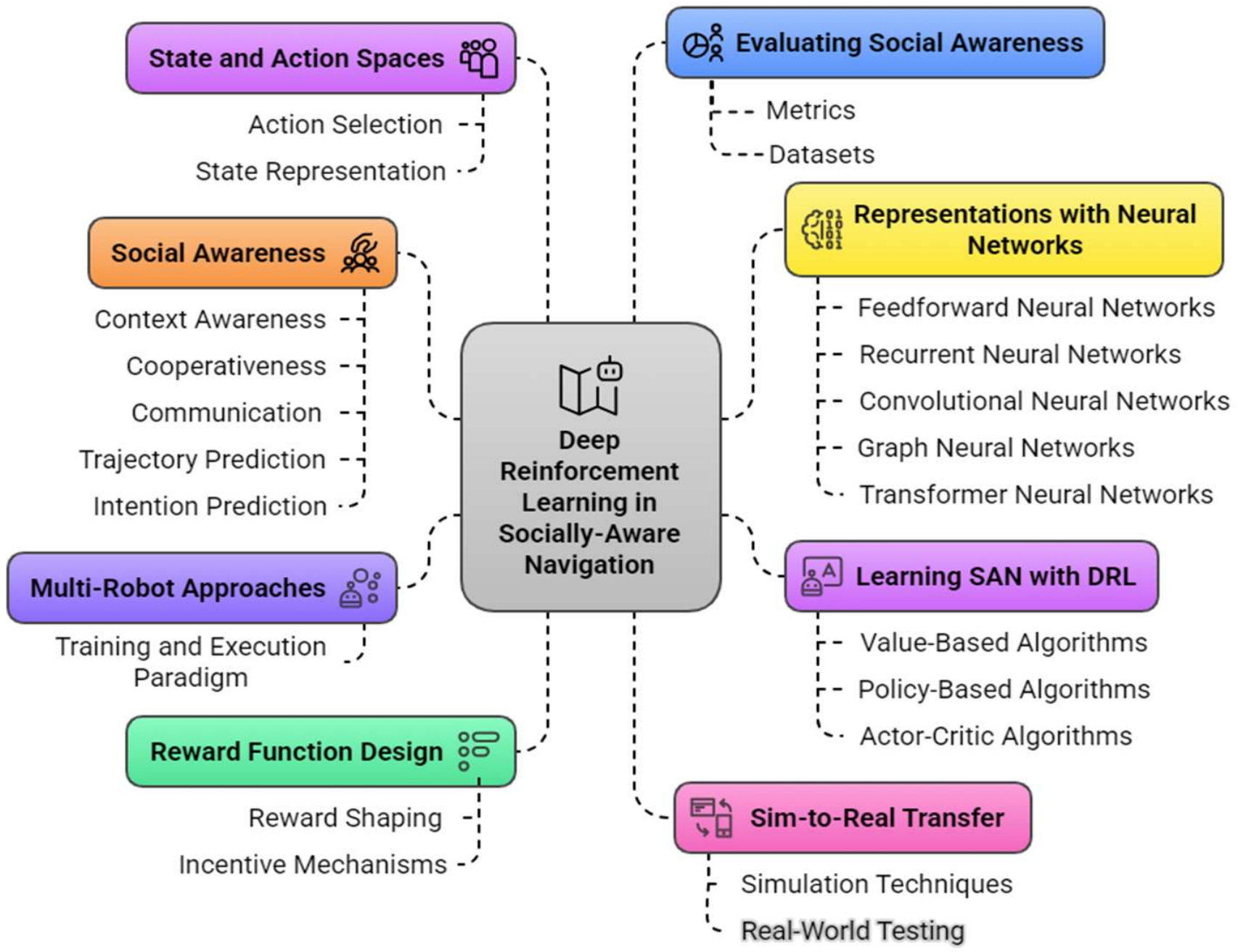

**Fig. 2** An overview of Concepts within DRL-based socially aware navigation

Several studies have conducted surveys in the research area of mobile robot navigation using deep reinforcement learning. However, few works have provided overviews on works relating to human-aware or socially aware robot navigation based on DRL, which is the aim of this paper. Kruse et al. [5] presented state-of-the-art human-aware navigation for wheeled mobile robots and categorized it based on comfort, naturalness, and sociability. Rios-Martinez et al. [7] presented a review on socially aware navigation focusing on proxemics to improve the robot's sociality. Another survey of close relevance, Multi-Agent Deep Reinforcement Learning (MADRL) for Multi-Robot applications, was presented by Orr and Dutta [8]. It explores the different types of multi-agent algorithms with a specific focus on multi-robot approaches. However, it does not discuss socially aware navigation. Singamaneni et al. [9] surveyed socially aware robot navigation and built a comprehensive and multi-faceted taxonomy. It delves into the topic from different angles, including robot type. It also expands on previous surveys by considering a broader range of robots classified as Ground (including legged and autonomous), Aerial, and Aquatic. It also discusses less-explored topics, such as human intention prediction, negotiation in navigation, and numerous evaluation metrics.

Akalin and Loutfi [10] present a survey on RL approaches in Social Robotics. They categorized the papers studied based on the type of RL framework, the communication medium used to formulate the reward function, the nature of the reward function, and the evaluation methodologies. The survey then proposed solutions to real-world RL problems, outlined potential issues, and identified open questions in the field. This paper, although closely related to ours, considers social robotics but does not discuss socially aware navigation, and provides an overview of the Reinforcement Learning approaches used in social robotics. Table 1 highlights how our survey paper compares with other recent and relevant survey works, and Figure 2 provides an overview of the different concepts considered in this work.

In [11], a comprehensive review of social navigation of mobile robots was done from the perspective of requirements, planning and evaluation of the navigation. It presented a taxonomy of the needs for perception and mechanism of planning as well as modalities for evaluating the socialness of the navigation. It then highlighted the main challenges and suggested perspectives for future work. The review presented in [12] surveys social navigation approaches for service robots, focusing on the establishment of human-robot interaction models, the application of subjective emotion state recognition in socially aware navigation, and the implementation of socially aware navigation under various learning frameworks.

While all the previous surveys have addressed various aspects of socially aware navigation, none have examined the joint integration of Deep Reinforcement Learning and Social awareness in navigation of mobile robots. Generally, previous surveys are limited to one of the three following categories:

1. Works that focus on traditional SAN with minimal exploration of learning paradigms.
2. Works that explore reinforcement learning applications in the broad field of social robotics without specifying navigation, or
3. Works that cover DRL only at the multi-agent level without integrating social awareness or Human-Robot Interaction principles in general.

This work, however, bridges these gaps through three key contributions:

1. It provides the first unified taxonomy that analyses and classifies the literature from both the algorithmic perspective (value-based DRL, policy-based DRL, and actor-critic based DRL) and neural network perspective (CNN, RNN, GNN, Transformer) thus illustrating how these design choices enhance socially acceptable navigation.
2. It performs a quantitative and qualitative comparison of evaluation mechanisms (metrics, datasets, simulators) revealing the lack of standardized social metrics and imbalance between technical and human-centred evaluation.
3. It synthesizes and compares over 100 DRL-based SAN works revealing insights, strengths, and emerging directions.

As highlighted in Table 1 no prior survey has simultaneously integrated DRL methodologies with socially aware navigation at this level of depth with the purpose of highlighting the design choices that go into enhancing social awareness of DRL-based SAN frameworks. Specifically, this work fills the gap between general DRL-enabled navigation and socially aware navigation, thus providing a comprehensive perspective that had not previously existed. To the best of our knowledge, no previous survey paper has done a comprehensive overview of the integration of social awareness to deep reinforcement learning-based mobile robot navigation. We believe this work will guide researchers and practitioners in selecting the most suitable and applicable DRL approaches and Neural network architectures for their specific requirements in encoding social awareness to navigation of mobile robots, and also highlight emerging research directions and unsolved problems in this research area.

**Table 1 Comparison of Review Papers related to SAN**

| **Title** | **Scope & Topics covered** | **Depth of DRL Integration** | **Social Awareness Factors** | **Novel Contributions** |
|---|---|---|---|---|
| From Proxemics Theory to Socially-Aware Navigation: A Survey [7] **(2015)** | Social Behaviour, Signals and Cues, Proxemics, SAN, and Interaction Types | No DRL coverage | Human comfort zones, proxemic zones, and levels, and social cues | First work to survey social conventions relevant to navigation |
| A survey on human-aware robot navigation [13] **(2021)** | Active Vision Paradigm, Visual robot navigation, HRI, Modelling Human behaviour | Minimal DRL integration | social interactions, proxemics, activity recognition, trajectory prediction, naturalness | Comprehensive human-aware navigation with an interdisciplinary perspective |
| Multi-Agent Deep Reinforcement Learning for Multi-Robot | DRL Theoretical Background and DRL Applications for MultiRobot Systems | Comprehensive and full emphasis on DRL | Social Awareness is not considered | Most recent review on MADRL techniques for MultiRobot Applications |
| Applications: A Survey [8] **(2023)** | | | | |
| A Survey on Socially Aware Robot Navigation: Taxonomy and Future Challenges [9] **(2024)** | Taxons for Robot type, Planning and Decision-Making, Situation Awareness, and Evaluation. | Mentions some DRL works | Social Norms and Proxemics. Trajectory and Intention prediction, and social cues. | Multi-faceted and comprehensive taxonomies integrating multiple disciplines and unique approaches |
| Bridging Requirements, Planning, and Evaluation: A Review of Social Robot Navigation [11] **(2024)** | Requirements-based taxonomy; perception, planning, evaluation; mapping algorithms to social requirements | Moderate.D RL is discussed as one planning family. | Safety, proxemics, legibility, approach speed/direction, occlusion avoidance, smoothness, gaze, passing side, yielding, queuing, elevator etiquette | Taxonomy of social navigation needs in reference to perception, planning, and evaluation. Maps diverse algorithms to these requirements with benchmarks/metrics |
| Reinforcement Learning Approaches in Social Robotics [10] **(2021)** | Technical background on RL. RL approaches based on RL Type and Reward Function. Evaluations | Covers RL but not fully DRL | Communication, Attention, Comfort, Human Engagement, and Facial Expressions. | First structured survey of RL methods in social robotics |
| A Review on Social Awareness Navigation for Service Robots [12] **(2025)** | Human–robot trajectory modeling, emotion recognition, learning frameworks (supervised, RL, preference) | Moderate. Summarizes RL works. No focus on DRL. | Proxemics, trajectory compliance, emotion-driven adaptation (facial, speech, gait), and social interactions in trajectory forecasting | Integrates emotion recognition with navigation frameworks. Highlights preference learning as an emerging direction |
| This work | Integration of DRL with SAN, Role of types of DNNs and RL Algorithms, Evaluations | Complete integration with emphasis of the role of DRL in SAN | Trajectory Prediction, Intention Prediction, Proxemics, Comfort, Naturalness, HRI | Classification of approaches across RL types and NN architectures. Emphasizes how Deep Learning and RL are interwoven in defining socially compliant navigation behaviors |

This survey focuses on works from 2017 to date (2025), as most works were done within this period, with Figure 3 showing the rising popularity of this field. The papers were obtained from Search databases such as Google Scholar, IEEE Xplore, and Scopus search. The search terms used are ‘Socially-aware’ OR ‘Human-aware’ OR ‘crowd-aware’ AND ‘Deep reinforcement learning’ AND ‘Mobile Robot’.

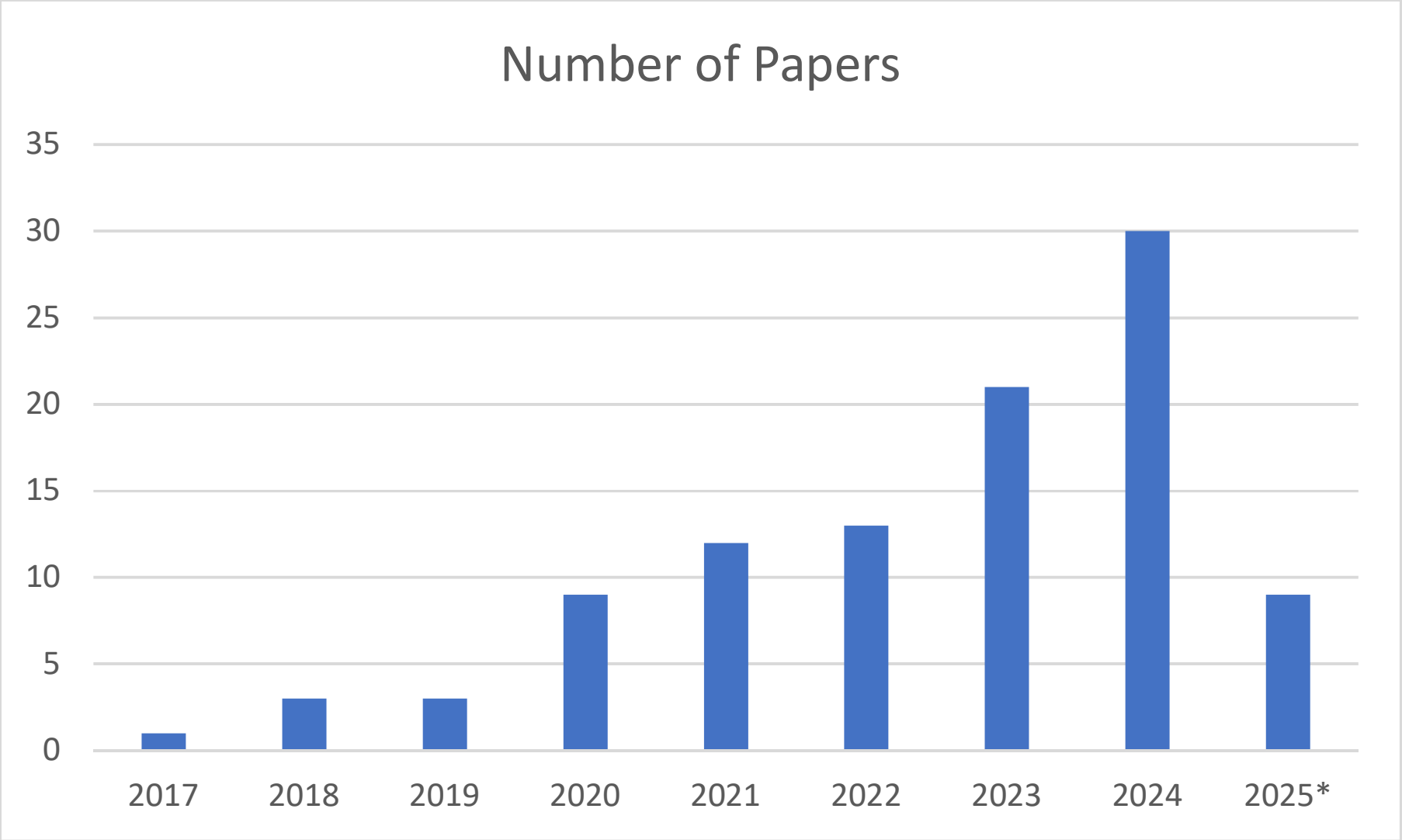

**Fig. 3** Paper publication in the past 9 years

The paper is organized as follows: In Section 2, we discuss and provide a background to Socially aware navigation including approaches, benefits and challenges of applying RL in the social robotics domain, along with a categorization and analysis of DRL approaches. In Section 3, we present the different Neural Network architectures and how they are suitably used in learning social behaviour and agent representation. Section 4 presents the mechanisms for evaluating social awareness of algorithms, and comparatively discusses the datasets and simulators applied to this process. Section 5 then discusses the challenges and future directions in this field. Finally, in Section 6 we conclude the paper and summarize relevant findings

## 2. LEARNING IN SOCIALLY AWARE NAVIGATION

### 2.1 Approaches in SAN

The first attempt to integrate social awareness into robot navigation systems was tour guide robots deployed in museums in the late 1990s, such as the RHINO [11] and MINERVA [12] robots in Germany and the US, respectively. These robots used localization and mapping to engage users in natural human settings [14]. The works on socially aware navigation can generally be classified into: reactive methods, proactive methods, and learning based methods as proposed by [15].

#### 2.1.1 Reactive methods

These methods react to other mobile agents in the environment and can be considered to be purely rule-based, without any prediction or learning happening. The social forces model was one of the earliest of these methods to be used in modeling socially aware navigation [16]. This model represents movements using social forces, which can be attractive or repulsive when navigating to a goal or avoiding an obstacle. However, these methods are severely limited in capturing complex interactions when applied in crowded environments [17]. The Velocity Obstacles was developed in [18] and has been frequently implemented in traditional robot navigation, as well as in socially aware navigation works in general. Another technique, Reciprocal Velocity Obstacles, proposed in [19], fundamentally assumes that each agent navigates independently and without communication with other agents, and they have similar (reciprocal) collision avoidance behaviours. However, it does not effectively handle static obstacles or fully capture more complex human interactions.

The Optimal Reciprocal Collision Avoidance (ORCA), another velocity-based technique for reciprocal n-body optimal collision avoidance, is proposed in [20]. It allows multiple robots to navigate independently and safely in complex environments without explicit communication. However, this technique suffers significantly when agents move at constant or near-constant speeds.

### 2.1.2 Proactive methods

These methods assume that humans cooperate towards navigation and do not behave in adversarial manners. These methods are also called trajectory-based methods because they must predict the behaviour of humans through their trajectories and plan a path to follow. The trajectories or behaviours are obtained from large-scale datasets to form a Prediction Motion Model, and then a Path Planning method is incorporated. By proactively considering human actions, these methods achieve smoother interactions, which lead to robot navigation within human social norms. An example of this class of approaches is the Interacting Gaussian Process (IGP) Method, proposed in [21], which uses the prediction of future states of robots and Gaussian Processes to estimate crowd interaction data and model socially acceptable navigation. However, it may suffer from balancing between optimality and computational complexity. Another method involves predicting pedestrian mobility using Monte Carlo Tree Search (MCTS) and Y-net. This strategy enhances the robot's navigation decisions in congested areas by simulating future states and estimating local goal quality [22].

### 2.1.3 Learning-based methods

In more recent years, learning-based methods have been developed and have become widespread. Such methods include Models that are based on Deep learning, Deep Reinforcement Learning, and Inverse Reinforcement learning [23], [24]. Deep Reinforcement Learning (DRL) has become widely used in mobile robot navigation to learn optimal navigation strategies incorporating social awareness. The work in [25] was one of the earliest to incorporate social awareness into collision avoidance with DRL. The Long-Short Term Memory (LSTM) Neural Network architecture was used in [26] to encode other agents and model the collective impact of the crowd in a socially aware DRL framework.

The Social Attention Reinforcement Learning (SARL) approach was proposed in [27]. It explicitly models the interactions between the robot and the crowd and uses self-attention technique to find the aggregate impact of the crowd. Numerous other learning based methods have been developed which apply Long-Short Term Memory (LSTM) [26], Self-Attention Mechanism [27], Graph-based Convolutional Neural Networks [28], [29], and Spatial-Temporal Transformers [30] in different scenarios to learn agent representations.

Imitation learning has also been used to obtain a navigation model based on the training data. In [31], Generative Adversarial Imitation Learning (GAIL) was used with visual inputs to obtain a socially aware method for movement of mobile robots. Imitation Learning has also been combined with RL for single goal navigation in maples scenarios [32].

## 2.2 Benefits and Evolution of DRL for SAN

The progress of research into DRL has enhanced mobile robot navigation by equipping robots with better capabilities to navigate in complex human environments, particularly those that require the application of social norms and proxemics. Deep Reinforcement Learning enhances socially aware navigation techniques by significantly improving some of its main limitations. Traditional techniques often assume humans are static obstacles that cannot adapt dynamically to human behaviour in real-time environments [33]. Traditional approaches also often rely on predetermined models, such as the Social Force Model and Optimal Reciprocal Collision Avoidance (ORCA) models, which cannot often adapt to the dynamic social behaviours of humans in the environment. DRL, however, introduces the ability to develop navigation policies that can learn from interactions with the environment to improve the robot's ability to navigate with enhanced socialness [34].

Initial research on reinforcement learning in robot navigation focused on obstacle avoidance without necessarily incorporating social behaviours or interactions into the reinforcement learning framework. Researchers then began exploring how RL could be applied to enhance the robot's ability to navigate in environments shared with people, considering factors such as proxemics, norms, and collaborative behaviour [35]. With the introduction of DRL, it became easier to create navigation policies that could learn from interactions with the environment and produce more human-like motion that adheres to social norms more effectively [34]. Recent advancements in DRL for socially aware navigation aim to solve the challenge of formulating the most appropriate reward function using multi-objective decision-making frameworks, developing time-efficient navigation policies, and developing innovative reward functions [36] and [37].s

### 2.3 Deep Reinforcement Learning

Deep Learning is the use of Deep neural networks as function approximators to handle complex and high-dimensional problems that traditional reinforcement learning techniques may not be able to deal with. The neural networks have multiple layers of neurons, hence the descriptive "deep". By utilizing neural networks, DRL enables the approximation of intricate functions and policies, thereby enhancing the agent's ability to learn from high-dimensional sensory inputs and to adapt its behavior based on interactions with the environment.

Reinforcement Learning is a type of Machine Learning where an algorithm tries to learn solutions for sequential decision processes via repeated interaction with the environment [38]. At every time step, the agent chooses an action based on an observation from the environment. The environment then changes its state in response to the action taken based on predefined transition dynamics and sends a reward signal to the agent. This process is illustrated in Figure 4. Hence, the outcome of this process is an optimal policy that selects actions in each state to maximize the expected return. However, this learning process is characterized by the exploration-exploitation problem, which aims to balance exploring the outcomes of different actions against staying with actions that have currently been the best. Deep Reinforcement Learning combines these two concepts of Deep Learning and reinforcement learning to leverage the strengths of both methodologies, enabling agents to learn from high-dimensional data while simultaneously optimizing their decision-making processes through interacting with the environment.

Reinforcement Learning can be formalized using Markov Decision Process (MDP) which serves as a conceptual framework for representing decision situations in which outcomes are affected by a series of decisions and their random events. The MDP describes a decision maker (agent) in a state, $s$, taking an action $a$, which based on probability changes the state and gets a reward $R$. In this new state, it must make a choice on which action to subsequently execute, and this continues until a stopping criteria is met. The MDP can be described by a tuple $< S, A, P, R, \gamma >$ where–s belongs to a set of states, $S$, called State Space, $a$ also belongs to a set of actions, $A$, called Action Space, $P$ is the probability that an action $a_t$ in a state $s_t$, will cause a transition to state $s_{t+1}$, . This is called the transition probability, and is given by $P = \mathbb{P}(s_{t+1}|s_t, a_t)$. $R$ is the scalar value of the reward received after the change from $s_t$to $s_{t+1}$ while taking the action $a_{t.}$ A discount factor, $\gamma$ is introduced to assess the impact of future rewards when calculating the total return of a current decision.

The expected cumulative reward for each state is the value $V_\pi(s)$, and the optimal policy $\pi^*$ that provides the maximum expected reward for each state is

$$\pi^* = \underset{a}{\operatorname{argmax}} V^*(s) \quad (1)$$

The term $Q_\pi(s, a)$ can also be called the Q-function. It is the value of an optimal policy when the first selected action is imposed. The optimal Q-function can be obtained by using a model-free reinforcement learning algorithm called the Q-learning algorithm. It estimates the maximum value of the value function of each action, and the optimal strategy can be determined by iteratively finding the extreme value of the Q-function directly. In Algorithm 1, we present its pseudo code sequentially. The exploration rate, $\varepsilon$ is used in determining the frequency of taking unknown actions during the learning process. The process function is highlighted in Algorithm 1.

**Algorithm 1** Q-learning

**Input:** the policy $\pi$

**Initialization:** $Q(s,a)$ Using arbitrary values (e.g. zeros) initialize Q(s,a)

**for** each episode **do**

initialize $s$

**for** each non-terminal state $s$ in each episode **do**

$a \leftarrow$ action for $s$ derived by $Q$

select action $a$, observe $r$, $s'$

$Q(s,a) \leftarrow Q(s,a) + \alpha[r + \gamma \max_{a'} Q(s',a') - Q(s,a)]$

$s \leftarrow s'$

**end for**

**end for**

**Output:** action-value function Q

---

### 2.4 Classification of DRL-based SAN Algorithms

Most works based on socially aware Deep reinforcement algorithms can be classified into value-based, policy-based, and actor-critic algorithms.

#### 2.4.1 Value-based RL Approaches

In RL, Value-based methods involve an agent with a function that takes in the state and one possible action and outputs the value of taking that action. The value is the total expected reward from a state up to future states. The policy of the algorithm is to check the value of every possible action and choose the one with the highest value. This can be represented using a lookup table called Q-table. The algorithm uses the Bellman equation through Dynamic programming to update the value of actions. In Value-based DRL, neural networks are introduced to represent the value functions. The value-based DRL obtains the agent's policy by continuously updating the value function [39].

Value-based DRL methods have been the most popular methods implemented in DRL for socially aware navigation. In one of the foundational works of this research area, [23] implemented a Value Network trained using the Temporal Difference method with Standard Experience replay. Several other subsequent works have implemented the same Deep V-Learning for training [25], [34], [40], [41].

Value-based approaches are frequently used because they directly assign values to actions, which can lead to more deterministic and stable decision-making processes. This direct approach can be particularly beneficial in environments requiring precise and consistent behaviour, such as socially aware navigation, where safety and compliance are critical. Value function-based methods also improve learning efficiency by focusing on the value of actions rather than the policy itself, which leads to faster convergence and more reliable performance in complex environments [36]. Value-based methods are more suitable for application in simple navigation scenarios with few agents and likewise few actions available for the robot to take at every point, for example, moving left, right, forward, backwards, or waiting. Value-based DRL methods also perform well in scenarios with sparse reward settings, where interactions between agents are few, so computational resources are minimized. This makes them particularly suitable for deployment on physical robots to perform real-world socially aware navigation. Although value function-based methods offer numerous advantages, they face limitations and challenges. For instance, as the learning environment grows, the computational demands of value-based methods can increase significantly, potentially leading to performance degradation.

The Deep Q-learning method is popular among the different types of value-based methods that have been applied to socially aware navigation. It involves extending the off-policy Q-learning algorithm with neural networks. This extension leads to the formulation of the deep Q-network (DQN) [42]. DQN is a standard building block for many single- and multi-agent RL algorithms [38]. It defines a neural network Q to represent the action-value function for a deep version of Q-learning. It is well-suited for handling discrete action spaces, i.e., in navigation, where the robot has a finite number of actions. This may be advantageous to navigation as it increases the comfort of humans because they can predict which action the robot may take, especially if they have had previous interactions with the robot. It is computationally efficient because it computes the action-value estimates for all actions in one forward pass through the network. Additionally, it helps the robot learn from past experiences stored through the experience replay mechanism. However, deep Q-learning is limited to environments with finite (discrete) action spaces like tabular Q-learning. In [43], a new DQN algorithm was proposed based on [44] that introduces an LSTM neural network into the value network. DQN was also implemented in [45], and then in [46], an extension of it was presented, called the Safety-Critical Deep Q-Network (SafeDQN). The authors built a computational model of patient acuity to generate the reward for the RL agent. However, DQN exacerbates the moving target problem when representing the value function with function approximation and tends to cause the network to overfit in the most recent experiences [47]. A comparison of different Value-based SAN approaches is provided in Table 2.

To enable systematic comparison between the different algorithm types and across the types of Neural networks, we have formulated Tables 2 to 4 which give a comparative analysis across some key social factors. These dimensions include:

1. Comfort: Human perception of safety and ease during robot interaction
2. Naturalness: The extent to which robot motion resembles human-like navigation behavior
3. Trajectory prediction: The robot's ability to anticipate human motion for collision-free and legible navigation
4. Attention mechanism: Whether the algorithm uses a mechanism to prioritizes relevant social or spatial cues
5. Physical input: The sensory modalities through which the robot perceives its environment
6. Context awareness: The ability to adapt motion based on environmental and social context
7. Intention prediction: The capacity to infer human goals or motion intent for proactive responses
8. Advanced socialness: Higher-level integration of emotional, psychological, or multimodal social cues beyond motion-based reasoning.

**Table 2. Comparative Analysis of Some Value-Based Approaches**

| Ref | Algorithm Type | Type of DNN | State and Action Space | Single/ Multi-robot | Testing Framework | Comfort | Naturalness | Trajectory Prediction | Attention Mechanism | Physical Input | Context Awareness | Intention Prediction | Advanced Socialness |
|---|---|---|---|---|---|---|---|---|---|---|---|---|---|
| **[48]** | Q-learning | FFNN | CC | Single | Sim | ✓ | - | ✓ | ✓ | - | - | ✓ | - |
| **[49]** | Q-learning | FFNN | CD | Multi | Sim | ✓ | - | - | - | ✓ | - | - | - |
| **[50]** | DQN | CNN | CD | Multi | Sim | ✓ | - | - | - | ✓ | ✓ | - | - |
| **[30]** | DQN | Transformer | CD | Single | Sim | ✓ | ✓ | - | ✓ | - | - | - | - |
| **[36]** | DQN | FFNN | CD | Single | Sim | ✓ | - | ✓ | ✓ | ✓ | ✓ | - | - |
| **[43]** | DQN | LSTM | CD | Single | Sim | ✓ | - | - | - | ✓ | - | - | - |
| **[51]** | Q-learning | LSTM | CD | Single | Sim | ✓ | - | - | - | - | ✓ | - | ✓ |
| **[52]** | DQN | CNN | CD | Single | Both | ✓ | - | - | - | ✓ | - | - | - |
| **[53]** | DQN | FFNN | CD | Single | Sim | ✓ | - | - | ✓ | - | - | - | - |
| **[54]** | DQN | GNN | CD | Single | Sim | ✓ | ✓ | - | ✓ | - | - | - | - |
| **[55]** | DQN | FFNN | CC | Single | Both | ✓ | ✓ | - | - | ✓ | - | - | - |
| **[56]** | DQN | FFNN | CD | Single | Sim | ✓ | - | - | - | - | ✓ | - | - |
| **[57]** | DQN | Transformer | CD | Single | Sim | ✓ | - | - | ✓ | - | - | - | - |
| **[58]** | DQN | CNN | CC | Single | Both | ✓ | ✓ | ✓ | - | - | - | - | - |
| **[59]** | DQN | FFNN | CD | Single | Sim | ✓ | - | - | - | ✓ | ✓ | - | - |
| **[60]** | DQN | CNN | CD | Single | Both | ✓ | - | - | ✓ | - | - | - | ✓ |
| **[61]** | DQN | Transformer | CC | Single | Sim | ✓ | - | - | - | - | ✓ | - | ✓ |
| **[46]** | DQN | CNN | CD | Single | Sim | ✓ | - | - | - | - | ✓ | - | - |
| **[34]** | DQN | FFNN | CC | Single | Sim | ✓ | ✓ | - | - | - | - | - | - |
| **[40]** | DQN | Transformer | CD | Single | Sim | ✓ | - | ✓ | ✓ | - | - | ✓ | - |
| **[62]** | DQN | GNN | CD | Single | Both | ✓ | - | ✓ | ✓ | - | - | ✓ | |
| **[63]** | DQN | CNN | CD | Single | Sim | ✓ | - | - | - | ✓ | - | - | - |
| **[64]** | Deep V-learning | FFNN | CD | Single | Both | ✓ | - | - | - | - | - | - | ✓ |
| **[65]** | Q-learning | GNN | CD | Single | Both | ✓ | ✓ | - | ✓ | ✓ | - | - | - |
| **[66]** | DDQN | GNN | CC | Single | Both | ✓ | ✓ | - | ✓ | ✓ | ✓ | - | ✓ |

Continuous and Continuous- CC; Continuous and Discrete- CD; Sim- Simulation; Both – Physical Robot and Simulation, Graph Neural Network- GNN; Feedforward Neural Network- FFNN; Convolutional Neural Network – CNN; Recurrent Neural Network- RNN; Long Short-Term Memory

### 2.4.2 Policy-based RL approaches

These methods may also be referred to as policy-gradient-based algorithms. These methods directly learn a policy that maps states to actions without relying on learning a value function. This policy is represented by any function approximation technique or, in the case of DRL, by neural networks. This provides more flexibility in selecting actions compared to Value-based RL algorithms and makes it possible to represent policies for continuous action spaces [38]. As a result of Policy-based algorithms' capabilities in effectively handling continuous action spaces, the robot can take more fine-tuned and nuanced actions involving changes in speed, steering angle, or acceleration when required. This fine-tuned motion control is vital in crowded and dynamic environments because it produces more human-like and natural motion than sudden jerky motions, which may arise when using Value-based methods. Some Policy-based DRL algorithms that have been applied in socially aware navigation include the Proximal Policy Optimization algorithm (PPO) [67], Deterministic Policy Gradient (DPG) [68], and Trust Region Policy Optimization (TRPO) [69].

The Proximal Policy Optimization Method is an improvement on the TRPO and has found good application in this field because of its more computationally efficient approach in trying to improve the quality of policies when introducing a penalty, as in the TRPO. It computes a more efficient objective to avoid large jumps in policies in a single optimization step. This is advantageous because, in complex and dynamic environments where robots navigate among humans, the learning has to be stable and reliable, given that humans can behave unpredictably, which may lead to robot behaviour that compromises human safety and comfort. The PPO has been used for policy and value function learning in [69] and [63]. Similarly, PPO was used in [70] to learn a critic network and a policy simultaneously. In [48], a Lagrangian method is implemented as a type of Safe RL technique to incorporate safety constraints into the PPO DRL framework. The PPO is widely used in the papers surveyed in this work due to its numerous benefits, including the ability to generalize across diverse social environments, handling complex multi-agent social interactions, and efficient handling of sparse and delayed rewards.

In [71], a Context-appropriate Social Navigation approach was developed for application in construction environments of various densities and incorporated into both the TRPO algorithm and the DQN algorithms to compare with [23], which proved to have better performance in high-density settings. These approaches are highlighted and compared in Table 3.

**Table 3 Comparative Analysis of Some Policy-Based Approaches**

| Ref | Algorithm Type | Type of DNN | State and Action Space | Single / Multi-robot | Testing Framework | Comfort | Naturalness | Trajectory Prediction | Attention Mechanism | Physical Input | Context Awareness | Intention Prediction | Advanced Socialness |
|---|---|---|---|---|---|---|---|---|---|---|---|---|---|
| [72] | PPO | GNN | CC | Single | Sim | ✓ | - | - | ✓ | ✓ | - | - | - |
| [73] | MAPPO | GNN | CC | Multi | Sim | ✓ | - | - | ✓ | ✓ | ✓ | - | ✓ |
| [74] | PPO | FFNN | CC | Single | Both | ✓ | - | - | - | ✓ | ✓ | - | - |
| [75] | PPO | CNN | CC | Single | Sim | ✓ | - | - | ✓ | - | - | ✓ | - |
| [76] | Model-based PPO | RNN | CC | Single | Both | ✓ | - | - | - | - | - | - | - |
| [77] | PPO | FFNN | CC | Single | Both | ✓ | - | - | - | ✓ | - | - | - |
| [78] | PPO | RNN | CC | Single | Sim | ✓ | - | - | ✓ | - | - | - | - |
| [79] | PPO | RNN | CC | Single | Sim | ✓ | - | - | ✓ | - | - | - | - |
| [80] | PPO | RNN | CC | Single | Sim | ✓ | - | - | ✓ | - | - | - | - |
| [26] | PPO | RNN | CC | Single | Both | ✓ | - | - | - | - | - | - | - |
| [81] | PPO | GNN | CC | Multi | Sim | ✓ | - | ✓ | ✓ | - | - | ✓ | - |
| [82] | PPO | Transformer | CC | Single | Sim | ✓ | - | - | ✓ | - | - | - | - |
| [83] | PPO | FFNN | CC | Single | Sim | ✓ | ✓ | - | - | - | - | - | - |
| [84] | PPO | GNN | CC | Single | Both | ✓ | ✓ | - | ✓ | - | - | - | - |
| [85] | PPO | Transformer | CC | Single | Both | ✓ | - | - | ✓ | - | - | - | - |
| [86] | PPO | GNN | CC | Single | Both | ✓ | - | - | - | - | - | - | - |
| [87] | PPO | Transformer | CC | Single | Sim | ✓ | ✓ | - | - | - | ✓ | - | - |
| [88] | PPO | GNN | CC | Single | Sim | ✓ | - | ✓ | ✓ | - | - | ✓ | - |

Continuous and Continuous- CC; Continuous and Discrete- CD; Sim- Simulation; Both – Physical Robot and Simulation; Graph Neural Network- GNN; Feedforward Neural Network- FFNN; Convolutional Neural Network – CNN; Recurrent Neural Network- RNN

### 2.4.3 Actor-critic-based RL approaches

Actor-critic methods, which are originally a family of policy-gradient algorithms, have been recently categorised by researchers and practitioners as a separate DRL approach because they combine elements of both value-based and policy-based methods [89]. These algorithms train a policy called the actor and a value function called the critic concurrently. The actor-network selects the best action based on the current state, and the critic network estimates the value of the actor's state and actions. Actor-critic networks inherently thrive in balancing exploration and exploitation when training the DRL policy. They can easily explore new behaviours towards the navigation goal while using the critic network to refine the policy based on feedback or reward signals. Additionally, they are well-suited to handle continuous high-dimensional state and action spaces, which are required in the most advanced SAN algorithms to learn advanced techniques like context-awareness, intention prediction, and trajectory prediction [90], [91]. The most popular algorithms under this include Advantage Actor-Critic (A2C) [92], Asynchronous Advantage Actor–Critic (A3C), Deep Deterministic Policy Gradient (DDPG)[93], Soft Actor-Critic (SAC)[94], Twin Delayed Deep Deterministic Policy Gradients (TD3) [95]. Table 4 provides a comparative analysis of the different works implementing these algorithms, focusing on socially aware navigation.

In [77], an A3C-based DRL framework was used to train a mobile robot to navigate in a socially aware manner by extracting the socio-spatial temporal characteristics of humans and projecting them on a 2D laser plane. An actor-critic approach is likewise implemented in [96] to train a socially aware algorithm that uses Graph Convolutional Networks to extract and represent human-robot interactions. However, actor-critic-based SAN algorithms struggle in dynamic multi-agent scenarios and have high computational costs due to simultaneously training actor and critic networks. Actor-critic methods are generally well suited for complex environments and multi-agent applications, and in Table 5, It is compared against the other two RL types for applications of SAN.

**Table 4 Comparative Analysis of some Actor-Critic-based approaches**

| Ref | Algorithm Type | Type of DNN | State and Action Space | Single / Multi-robot | Testing Framework | Comfort | Naturalness | Trajectory Prediction | Attention Mechanism | Physical Input | Context Awareness | Intention Prediction | Advanced Socialness |
|---|---|---|---|---|---|---|---|---|---|---|---|---|---|
| **[90]** | Multi-Agent Actor-Critic | GNN | CC | Multi | Sim | ✓ | - | ✓ | ✓ | - | ✓ | - | - |
| **[97]** | Soft Actor-Critic | GNN | CC | Single | Both | ✓ | - | - | ✓ | - | ✓ | - | - |
| **[98]** | Actor-Critic | FFNN | CC | Single | Sim | ✓ | - | - | - | - | ✓ | - | - |
| **[71]** | Actor-Critic | FFNN | CC | Single | Sim | ✓ | - | - | - | - | ✓ | - | - |
| **[99]** | Actor-Critic | RNN | CC | Single | Both | ✓ | - | - | - | ✓ | - | - | - |
| **[100]** | A3C | CNN | CD | Single | Sim | ✓ | - | - | - | ✓ | - | - | - |
| **[101]** | Actor-Critic | CNN | CC | Single | Both | ✓ | - | - | ✓ | - | - | - | - |
| **[102]** | Actor-Critic | FFNN | CC | Single | Both | ✓ | - | - | - | ✓ | - | ✓ | - |
| **[107]** | Actor-Critic | CNN | CC | Single | Sim | ✓ | ✓ | - | - | - | - | - | - |
| **[103]** | Actor-Critic | FFNN | CC | Single | Both | ✓ | - | - | - | ✓ | - | - | - |
| **[104]** | Actor critic | CNN | CC | Single | Both | ✓ | ✓ | - | - | ✓ | - | - | - |
| **[105]** | Actor-critic | FFNN | CC | Single | Sim | ✓ | - | - | - | - | - | - | ✓ |
| **[106]** | Actor-Critic | LSTM | CC | Single | Sim | ✓ | - | - | ✓ | - | - | - | - |
| **[107]** | Actor critic | GNN | CC | Multi | Sim | ✓ | - | ✓ | ✓ | - | - | ✓ | ✓ |
| **[121]** | Actor-Critic | Transformer | CC | Multi | Sim | ✓ | - | ✓ | ✓ | - | ✓ | - | ✓ |
| **[108]** | Actor-Critic | FFNN | CC | Single | Sim | ✓ | ✓ | - | ✓ | - | - | - | - |
| **[109]** | Soft Actor-Critic | GNN | CC | Single | Both | ✓ | ✓ | ✓ | ✓ | ✓ | ✓ | - | - |
| **[110]** | Hybrid PPO | GNN | CC | Single | Sim | ✓ | ✓ | - | - | - | - | - | - |
| **[111]** | Actor-Critic | GNN | CC | Single | Both | ✓ | ✓ | - | - | ✓ | - | - | - |
| **[112]** | A2C | CNN | CC | Single | Both | ✓ | - | - | - | ✓ | ✓ | - | ✓ |
| **[113]** | Soft Actor-Critic | CNN | CC | Single | Sim | - | - | - | - | - | - | - | - |
| **[114]** | Actor-Critic | GNN | CC | Single | Both | ✓ | ✓ | ✓ | - | ✓ | - | ✓ | - |

Continuous and Continuous- CC; Continuous and Discrete- CD; Sim- Simulation; Both – Physical Robot and Simulation, Graph Neural Network- GNN; Feedforward Neural Network- FFNN; Convolutional Neural Network – CNN; Recurrent Neural Network- RNN; Long Short-Term Memory

From the comparative analysis in Tables 2 to 4, we can extract several key insights. Most of the earliest works in DRL-based SAN primarily focus on comfort and safety through collision avoidance, typically relying on low-dimensional physical inputs and simple feedforward architectures. Over time, research developed into context-aware and intention prediction models that utilize richer sensory inputs (e.g., RGB or pose information) and attention-based or graph-based neural networks. This trend marks a shift from reactive to predictive social behaviors. Even though recent works more commonly integrate trajectory prediction and context awareness, advanced socialness features such as emotion inference, multimodal reasoning, or explicit modelling of human intentions remain relatively underexplored, which presents an emerging direction for future research. Another noticeable point is that simpler Neural network structures, such as FFNN, are commonly used for basic social features; however, incorporating more advanced social mechanisms usually involves more complex network architectures.

**Table 5 Comparison of RL Methods for Socially Aware Navigation**

| Type of RL | Strength | Weakness |
|---|---|---|
| **Value-Based** | Handles simple social behaviours like avoiding collisions, causing discomfort better and faster | -Struggles in handling complex social interactions<br>-Produces Rigid jerky behaviour, which can cause discomfort or a lack of safety |
| **Policy Based** | -Can better handle more complex interactions (yielding, waiting, following)<br>-Produces more natural motion<br>-Adapts better to dynamic social context and uncertainty | -Struggles in navigating effectively in dense crowds due to constant policy updates because of changing behaviours |
| **Actor-Critic** | -Effectively balances the safety and flexibility of robot navigation<br>-Works well in complex scenarios involving multiple agents | -High computational cost, which can cause delays in decision-making and, hence, discomfort to humans |

Analysis of the reviewed studies reveals key trends in the evolution in DRL approaches for socially aware navigation. While value-based methods such as DQN and its variants have the advantage of algorithmic simplicity and stable convergence, they often struggle with continuous control and social adaptability. Policy-based algorithms, offer smoother and more natural motion generation and are better suited for high-dimensional, continuous navigation spaces. However, they computationally expensive and require careful reward shaping to maintain stability. Actor–critic frameworks, which combine the strengths of both paradigms, show the greatest promise for real-world deployment. Their dual-network structure allows balancinfg of policy exploration with value estimation, enabling faster learning and improved generalization in dynamic, human environments. Actor–critic models also integrate more naturally with modern neural network architectures (e.g., graph networks and transformers) that encode relational and contextual information.

### 2.5 Reward Function Design for Socially Aware Navigation

Reward function design is critical to ensuring socially aware navigation using DRL-based methods because it guides the accurate modelling and definition of navigation rules in environments where humans and robots interact and navigate. The navigation efficiency depends on how accurate and well-designed the reward function is in capturing all the scenarios and key considerations for a robot to navigate in the given environment. This section provides an overview of the different reward and penalty mechanisms used in DRL and their impact on social awareness.

Collision penalties, Arrival rewards, and Uncomfortable distance penalties are fundamental because socially aware navigation, by definition, cannot be achieved without incorporating these, which is why all works covered by this survey include these. Collision penalties penalize (give a negative reward) when the robot collides with an object. Some works went further on refining collision penalties by differentiating between the type of obstacle and hence associating different penalty values based on whether they are static, dynamic, like pedestrians, and even distinct categories of humans [59], [74], [100], [115]. Arrival rewards provide a reward value when the robot has achieved its navigation goal, and uncomfortable distance penalties penalize a robot coming too close to a human, i.e., having a distance less than a minimum predefined comfortable distance. Some studies introduce step rewards [45], [80], [116], [117], which provide a reward to encourage the robot to continuously take actions towards the navigation goal to promote efficiency in taking actions.

Another vital reward mechanism is Time-based penalties, which penalize the robot for exceeding a time limit to reach the goal [49], [58], [108], [118]. These penalties encourage efficient behaviour but may not be advantageous in environments where safety and following social norms are prioritized, such as in dense, crowded spaces. Other social mechanisms include a penalty for violating social norms, such as moving against the flow of pedestrians [119] or crossing interaction spaces [100]. Modelling group dynamics plays an essential role in works that consider socially aware navigation of pedestrian groups, often called group-aware navigation. Social group-based rewards and penalties have been formulated, such as Penalties for breaking group cohesion or intruding into a group's space [120], [121], [122], rewards for helping a group reach their goal, and rewards for a group staying together [121].

Other mechanisms include Zone-entering penalties, which penalize the robot for entering specific danger zones or discouraged zones, which could be a radius around the human which, if crossed, causes discomfort or apprehension or a shaped radial sector which varies based on the activity, motion, or attention of the human [36], [62], [74], [77], [79], [102], [123], [124]. The advantage is that these spatial constraints can improve safety; however, being too restrictive may lead to inefficient navigation. Another unique mechanism is the use of a human choice reward, which provides a reward based on the variety of human states after multiple transitions [35] to encourage the robot to consider human reactions over time in selecting actions.

Feature-based rewards, such as those based on direction of approach or angle, refine the robot's behaviour by rewarding smooth path angles, which encourage smoother and more natural movement [49], [52], [80], [82], [119], [125], [126]. However, prioritizing these may lead to overfitting towards particular navigation styles. Similarly, some works include path efficiency rewards which reward selecting the most efficient path [70], [79], [105], [127] or minimizing time and distance [52], [82], [108], [123]. Another common mechanism is the proximity to goal (progress) reward, which continuously rewards the robot as it moves closer to its target over time [6], [59], [76], [80], [84], [102], [103], [127]. More advanced techniques include reward prediction-based [83] and trajectory prediction-based [107] mechanisms.

### 2.6 Effect of Discreteness of State and Action Spaces

The type of state and action spaces in DRL-based SAN methods impacts the efficiency, adaptability, and safety of mobile robot navigation in human-populated environments. The choice between Continuous and Discrete leads to trade-offs in terms of computational complexity and learning efficiency. Continuous action spaces enable more fluid changes in parameters like steering angles, velocities, and accelerations, leading to smoother navigation of the robots, especially in dynamic environments where robots need to adapt to changes in human behaviour. This leads to navigation behaviour that is more acceptable to humans. However, they are computationally complex, and learning is slower to converge due to the large number of actions that need to be explored [54]. On the other hand, discrete action spaces make learning easier by reducing the complexity of the action space because there is a limited number of predefined actions. This leads to faster training, but may result in jerky motions of the robot.

Similarly, continuous state spaces provide more details and environmental representations that help in modelling the interactions more accurately; however, they bring about high-dimensional complexity. In contrast to this, discrete state spaces simplify the learning process and are easier to model, but this reduces precision and adaptability to the ever-changing human behaviours. The use of continuous state and action spaces generally leads to better performance of SAN frameworks, but they are more complex and have a higher computational burden. Discrete state and action spaces, though simpler and easier to train, limit the accuracy and effectiveness of the robot in navigating in socially acceptable manners. These trade-offs are summarized in Table 6 below.

**Table 6 Comparison of Continuous and Discrete State and Action Spaces**

| | **Continuous Action Space** | **Discrete Action Space** | **Continuous State Space** | **Discrete State Space** |
|---|---|---|---|---|
| Advantage | Smooth, fluid movements; fine- | Easier, faster training; limited action set | Rich environmental details; more | Simplified training; easier to model |
| | tuned control for human interaction | | context-aware behaviors | |
| Challenge | High computational cost; slower convergence | Jerky movements; less adaptive in dynamic environments | High-dimensional complexity; harder to process | Reduced precision; limited adaptation to subtle changes |

**2.7 Multi-Robot Approaches**

Recently, researchers have begun to extend single-robot SAN methods to multi-robot approaches, as it is often necessary for multiple robots to navigate simultaneously in the same environment, in the presence of humans, to collectively or independently achieve their navigation goals. However, multi-robot navigation has its own challenges. Due to a larger number of agents in the environment, collisions are more likely to happen, a significantly larger amount of computation is required, and coordination or cooperation between the robots in the environment must be ensured. Multirobot approaches in SAN include the ability to leverage collective information and distributed decision-making to improve navigation outcomes.

Given that multiple robots are navigating in the same environment. Each has a requirement to navigate in a socially acceptable manner, there is a need to explicitly define the type of relationship between the robots in the environment when it comes to implementing the learning and navigation policies and the kind of communication and cooperation when carrying out the navigation policies. When the RL algorithm has access to only the information observable by each agent (robot) for the training of the policy, it is referred to as decentralized training, but if each agent can also use and incorporate information from other robots in the training process, it is referred to as centralized training. Similarly, when executing the trained policies, if the action selection (execution) by the agent is based on previous observations by the robot, it is a decentralized execution.

Hence, there are generally three RL paradigms based on training and executing policies: Centralized Training and Execution (CTE), Decentralized Training and Execution (DTE), and Centralized Training with Decentralized Execution (CTDE). The CTE paradigm involves the sharing of information between agents in a centralized manner, where all robots communicate their local observations and/or policies, and all agents take actions based on the central policy. This paradigm is particularly useful where complex coordination of robots is needed because it leverages the joint-observation space of the environment. However, methods that employ this paradigm face challenges in deriving a single reward from the joint reward across all agents. Another significant problem is that with the increasing number of robots, the joint action space grows exponentially, and this drastically affects learning.

In the DTE paradigm, learning navigation policies and action-taking are performed independently, based solely on the information locally available to each robot. This paradigm enables the increase in the number of robots without concern for exponential growth in action spaces. This, however, means that robots can't leverage information available only to other robots to improve their own navigation policies. Additionally, because many robot policies are running at the same time, training policies may suffer from poor convergence because a change in the policy of one robot will necessitate other robots to adapt their policies similarly, and this makes learning unstable. In [128], the author presented a Heterogeneous Relational DRL method for Decentralized Multi-Robot Crowd Navigation, which was based on a DTE paradigm.

For the CTDE paradigm, the robot policies are trained using shared information that is available to all robots, and the robots take actions based on only the local information available to each robot. Thus, CTDE leverages the strengths of both CTE and DTE. The works presented in [73], [90], [107], [129], [130] all implement the CTDE paradigm.

These concepts and classifications of DRL approaches and algorithms lay the foundation for understanding the learning paradigms for SAN, specifically how the agents explore, learn from rewards, and update their policies. However, to fully optimize the use of DRL in SAN frameworks, it is essential to design the underlying neural network architectures as well as the reinforcement learning frameworks. The Deep Neural Network architecture, however, determines the quality of these learned behaviours and the effectiveness of the robot in perceiving the environment, understanding human motion, respecting social norms, and generating smooth trajectories that avoid collisions. In simpler terms, DRL determines how the robot learns, while the neural network determines what it learns and how effectively that learning translates into the desired social behaviors. Hence the choice of DNN design directly determines the performance of the DRL algorithms, which necessitates an in-depth analysis of the various architectures with respect to ensuring social awareness in navigation. The next section, therefore, focuses on the various DNN architecture types that are used in SAN frameworks, highlighting their respective strengths, limitations, and application areas.

## 3. NEURAL NETWORK ARCHITECTURES FOR LEARNING SAN

In deep reinforcement learning (DRL) methodologies for SAN, it is crucial to represent the deep learning aspect accurately. This is realized through neural networks, which serve as function approximators, helping in the representation of non-linear value functions across diverse states. These networks play a crucial role in facilitating agent navigation and their appropriate interaction within human-based environments. Through acquiring high-dimensional features of human–robot interactions and environmental dynamics, the networks allow the robots to learn generalized behaviours, including the desired socially-aware navigation patterns.

This section examines the various neural network architectures employed in DRL-based SAN, highlighting their strengths and limitations in modeling agent representations and learning socially compliant behaviors. A comparative analysis of the neural network architectures is also done with reference to their application areas, with the aim of guiding researchers in selecting the most appropriate architecture type given their navigation requirements and desired application.

### 3.1 Feedforward Neural Networks (Multi-Layer Perceptrons)

Feedforward neural networks, also known as multi-layer perceptrons (MLPs), are the most commonly used type of Deep neural network architecture in SAN. This is due to their simplicity and ease of implementation, which makes it easier to generalize across state and action spaces. Additionally, they are easily integrated with other neural network architectures to improve learning. However, due to their fully connected structure, they are inefficient for use in state spaces with high dimensionality, such as in socially aware navigation of mobile robots in dense crowd scenarios. Additionally, MLPs struggle with defining relationships based on space and time.

MLPs are commonly used as value function approximators, particularly in value-based algorithms such as DQN and Deep V-Learning. Their inherent simplicity allows for efficient mapping of state-action pair inputs to Q-values. They are ideal for simple tasks that require guaranteed stability and convergence. Additionally, they are commonly used to form hybrid neural network architectures along with other types of neural networks. They have been virtually used or integrated into most of the works this paper surveyed [34], [36], [48], [49], [53], [55].

### 3.2 Graph Neural Networks

Graph Neural Networks (GNNs) are a type of NN that takes graph-structured data as input and allows classification and regression of graphs and nodes. GNNs are composed of layers that operate on a graph whose structure remains static, but the features are updated in every layer of the network [131]. GNNs are particularly useful in socially aware DRL-based methods that model complex interactions between agents in dynamic environments. GNNs can be used to design models that capture the relationship between nodes in a graph, making them ideal for tasks involving human-robot interaction and crowd-aware navigation. GNNs have also found widespread usage for multi-robot DRL navigation frameworks [90], [107], [132].

GNNs are commonly used for multiagent and actor-critic algorithms because of their strength in explicitly encoding relational information among humans and robots which allows the critic policy to reason over structured interaction graphs. Research works that have used GNN as the Deep Learning Architecture include [116], [116], [128], [133], [134] with subtypes like Graph Convolutional Networks (GCNs) being applied in [35], [75], [96], [101], [119] to capture more relevant information from closest agents and Graph Attention Networks (GATs) in [85], [126] which use attention mechanism to assign different weights to different agents in the graph based on importance in the specific context. GNNs are also advantageous because they excel at learning how entities (humans, robots, obstacles) interact with one another. This is important in modelling group dynamics, social interactions, and crowd movements. However, GNNs can be computationally expensive, leading to slow convergence and requiring well-defined graphs to model interactions, which may limit their real-time applicability in highly dynamic environments. An Illustration of how agents are represented in a Graph-based Socially Aware DRL framework is shown in Figure 5.

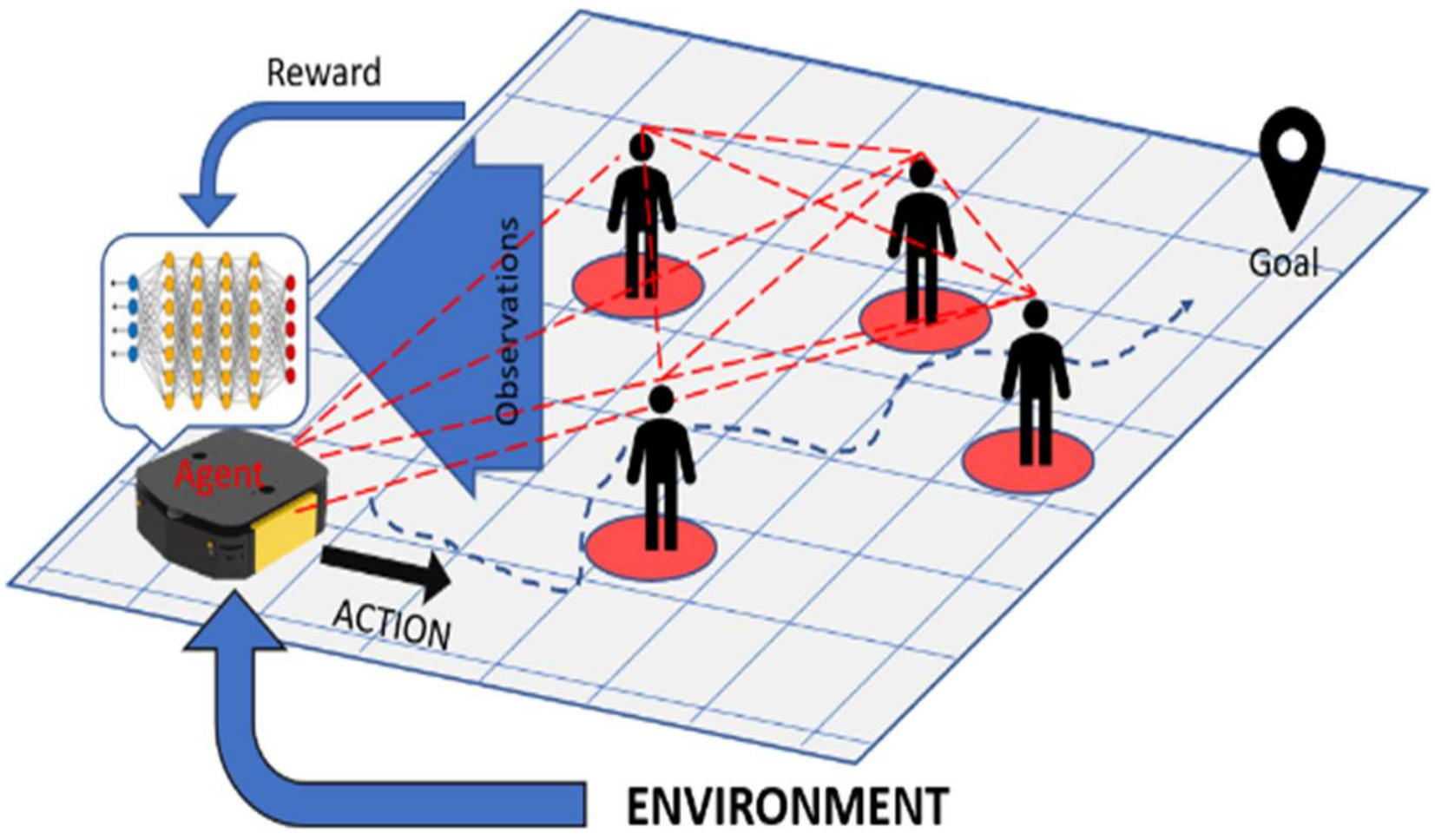

**Fig. 4** Illustration of a Socially Aware DRL framework with Graph-based representation

### 3.3 Recurrent Neural Networks (RNNs)

RNNs are neural networks that are employed for processing sequential data. RNNs are particularly suitable for use in DRL-based SAN frameworks because of their ability to model sequential dependencies and handle dynamic environments effectively. These capabilities are important for navigating dense human environments where it is necessary to model human behaviour and predict future positions and actions of people in the environment [79]. RNNs have been implemented in DRL methods in this field to predict how past actions and interactions influence future behaviours [6], [36], [99]. Other works that implement RNNs include [57], [78], [79], [91], [97], [107].

The most commonly used RNN architectures for SAN are the Long Short-Term Memory (LSTM) [135] and gated recurrent units (GRUs) [136]. LSTMs enhance traditional RNNs by solving the issue of vanishing gradients, allowing them to capture long-term dependencies between human actions and robot responses. LSTM models are more suitable for storing long-term social interaction data to process social norms and understand human behavior patterns accurately. LSTM-based models have been implemented in and [116] to model interactions in crowded and dynamic environments. GRUs are specifically designed to solve long-term dependencies [89]. GRUs are more suitable in cases where a robot needs to immediately respond to social input in the short term and in less dynamic social settings. GRU-based Neural Network architectures have been implemented in [106], [108], and [122] to enable quick action-taking for real-time navigation. While RNNs are effective for sequential tasks, their training is computationally intensive, and they can struggle with handling very long sequences or learning complex spatial-temporal interactions compared to more recent architectures like Transformers.

### 3.4 Convolutional Neural Networks

Convolutional neural networks (CNNs) are designed to process spatially structured data, such as images, by processing patches of nearby pixels at a time. This makes them highly effective for tasks involving visual perception in navigation. In socially aware navigation, CNNs are often used in cases where the robot must interpret visual data, such as detecting obstacles, human poses, or spatial layouts from camera input. CNNs use pooling operations to learn representations of high-dimensional inputs and reduce their dimensionality. This makes the network more robust to small changes in the input. CNN-based DRL architectures have been implemented in [50], [53], [58], [74], [101], [104], [125], [133] to enable the robots to process high-dimensional data such as camera images and LiDAR, and to make real-time navigation decisions which take context into account.

CNNs help identify and interpret human gestures, postures, and movements, which aid in predicting pedestrian trajectories and interactions with the robot [9]. By incorporating CNNs, DRL models can learn complex social interactions and respect social norms, such as maintaining personal space and avoiding collisions. CNNs can also be combined with other neural network architectures, such as graph neural networks (GNNs), to effectively model social interactions. However, CNNs struggle with handling sequential or relational data independently, which limits their application to tasks that require only spatial processing. For cases where spatial and temporal reasoning are needed, CNNs are often combined with other architectures like RNNs or LSTMs.

### 3.5 Transformer Networks

Transformer networks, which are relatively new, are a type of Deep learning model that can handle sequential data and long-range dependencies using mechanisms like self-attention and multi-head attention. Although similar to RNNs, transformers have the advantage of having no recurrent units, hence requiring less training time. Due to their effectiveness in processing complex temporal and spatial data, they can be used in pedestrian trajectory prediction and social navigation in dynamic environments [137]. Transformer-based networks have also been utilized to integrate spatiotemporal features into navigation policies, thereby providing superior performance in dynamic environments [138]. Transformer Networks have been implemented in [61], [62], [73], [87], [90], [130], where their advantages over RNNs are clearly showcased.

The clear advantage that Transformer networks have in handling both spatial and temporal data efficiently also poses a disadvantage because of the complex architecture and high computational cost. This can be problematic, especially in real-time applications, where a robot needs to make quick decisions when a sudden change in human behaviour occurs. To compare the advantages and disadvantages of the different types of Deep Neural Networks and their application areas, Table 7 is presented below.

**Table 7. Comparison of DNN Architectures with respect to SAN**

| Architecture Type | Application Area | Advantage | Disadvantage |
|---|---|---|---|
| **FFNN** | Simple Environments or in low-dimensional state spaces | Simple and efficient. Easy to train | It cannot model sequential data and is inefficient for high-dimensional inputs. |
| **GNN** | Capturing relational interactions among agents, social preference modelling | Works well with relational data and inter-agent relationships in multi-agent and dynamic scenarios | It is computationally expensive and requires expert knowledge. |
| **RNN** | Navigation that requires memory of past states and environments with sequential decision-making | Good with temporal dependencies and sequence prediction | Struggles with long-term relationship modelling, and it is computationally expensive. |
| **CNN** | Processing visual (camera) and LiDAR data. | Works well in spatial feature extraction and can handle high-dimensional image inputs. | Cannot handle temporal or sequential data directly |
| **Transformer** | Multi-modal inputs combining vision and language | Adaptable to multi-modal data and can model long-term dependencies and complex relationships | Requires massive datasets for effective training. Computationally expensive due to attention mechanisms. |

In conclusion, the integration of DRL algorithms and Neural Network Architectures defines how effectively a robot can perceive, reason and act within human environments. Value-based, policy-based, and actor–critic algorithms which rely on distinct neural networks which range from simple feedforward estimators to sophisticated graph and transformer models. By connecting the theoretical aspects of DRL with the architectural choices that lead to a well aligned integration and, this section establishes the foundation for evaluating and benchmarking socially aware navigation methods which is discussed in the next section of this paper.

## 4. EVALUATING SOCIAL AWARENESS OF ALGORITHMS

Evaluating socially aware navigation algorithms is fundamental in showcasing their advantages, performance, and scalability in more complex environments. This section provides an overview of the approaches used by the research community to comprehensively evaluate deep reinforcement learning-based SAN algorithms. The section discusses the metrics, Datasets, learning environments, simulators, and benchmarks used to evaluate the social awareness of DRL-based mobile robot navigation algorithms.

### 4.1 Metrics

It is necessary to select appropriate metrics for evaluating the algorithms to ensure that robots can navigate in human-rich environments in socially accepted manners. These metrics should provide a comprehensive evaluation of the navigation framework regarding both technical performance and social acceptability of the navigation behaviours. This is highly relevant in dynamic environments that may contain multiple agents (humans and robots), where the type of agents and interactions between them play critical roles.

Generally, evaluation methodologies encompass both qualitative and quantitative approaches. Qualitative methods are usually based on subjective evaluations, such as questionnaires administered during studies in which humans and robots have interacted, which assess humans' preferences and levels of comfort while engaging with the robots [139], [140]. These subjective assessments provide significant insights into the social acceptance of robotic navigation.

Quantitative methodologies use mathematically formulated metrics to evaluate both technical and social aspects of the robot's navigation. These methods enable accurate evaluation, allowing for justified comparisons among various navigational algorithms. The remainder of this subsection will focus on these methods.

The quantitative metrics can be classified into two categories: Navigational metrics, which assess the efficiency of the algorithms in terms of technical performance, and social metrics, which consider the social acceptability of the algorithms.

Some Navigational metrics include:

- **Success Rate:** This measures the frequency with which the robot successfully reaches its goal without collisions in a given time frame. This metric is fundamental and applicable to almost all socially aware navigation algorithms.
- **Collision Rate**: This metric measures the frequency of collisions of the robot with other agents or obstacles during navigation. This metric is also crucial in determining the algorithm's progress toward achieving the optimal policy.
- **Time Taken**: This defines the robot's time to reach its goal. It has been applied in [36], [43].
- **Path Length**: It measures the total distance the robot covers to reach its goal. Reducing the path length is considered an objective to enhance the navigational performance of the DRL socially aware algorithms. In [30], [141], this metric is considered.
- **Cumulative Reward**: Represents the total reward accumulated by the robot during an episode, showcasing the effectiveness of the reward function. This is another foundational metric in RL, as it helps to understand the algorithm's progress towards achieving an optimal policy.

Social Metrics include:

- **Comfort Metrics**: This category includes metrics like discomfort distance or discomfort frequency, which measure how often or how closely the robot approaches other agents, reflecting social awareness. This metric is essential in measuring how socially aware an algorithm is. It is overwhelmingly applied in most of the surveyed papers.
- **Intrusion time ratio**: This is the number of timesteps a robot takes to enter a human's future path. It measures how uncomfortable a robot makes a human. It has been considered in [40], [63], [78], and [142].
- **Social distance (during intrusion)**: This metric evaluates the robot's social compliance and navigation behaviour in proximity to humans. It is considered the mean distance between a robot and the nearest human at the instance of intrusion and has been widely applied in relevant papers, including [48], [81], [142], [143].

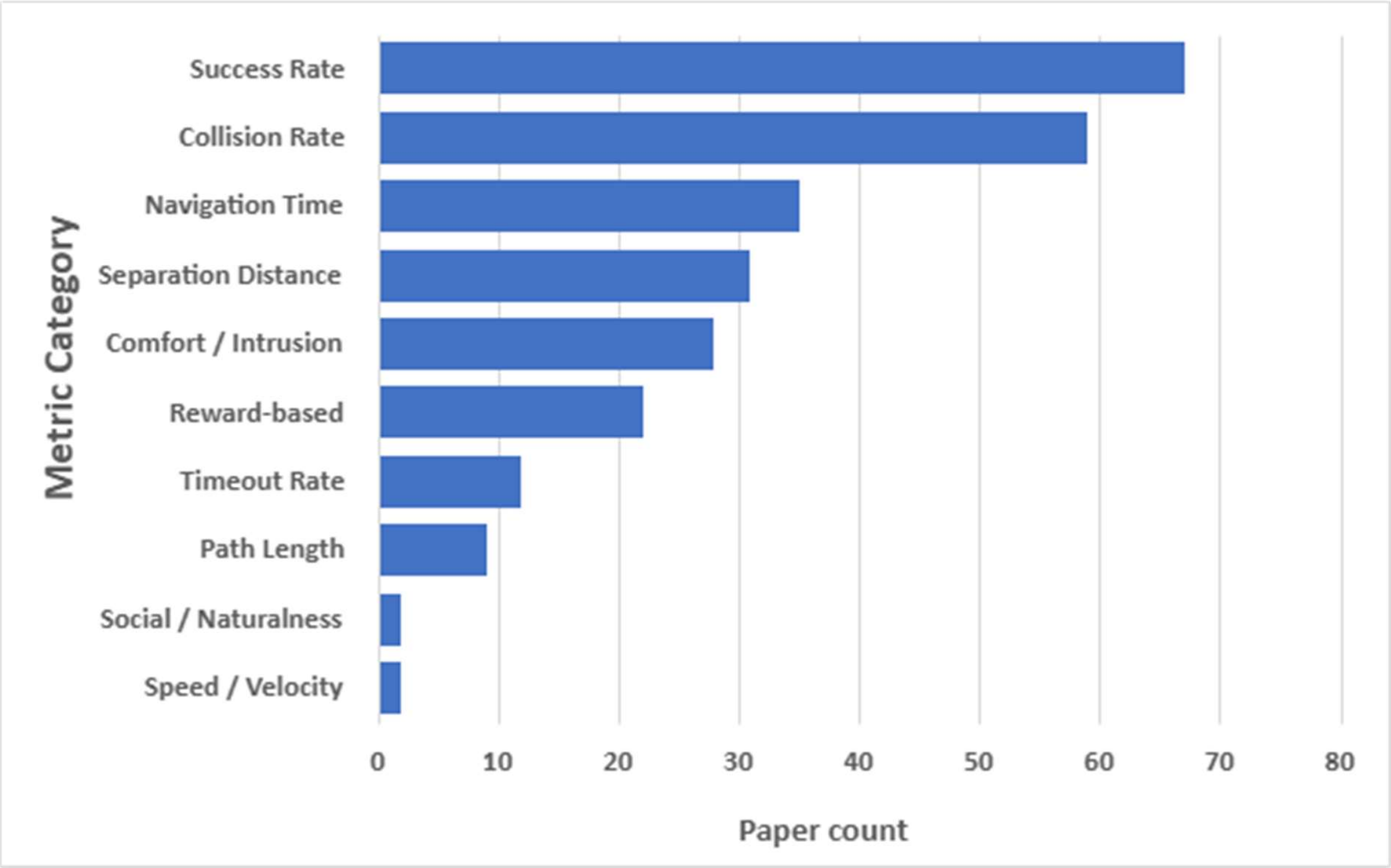

**Fig. 5** Analysis of Metrics applied in the surveyed papers

An analysis of the use of evaluation metrics in assessing the effectiveness of the algorithms showed a lack of uniformity in evaluation practices. Success rate (65%) and collision rate (57%) were the most common metrics used because they are fundamental in assessing navigation feasibility and safety. Navigation time (34%), Timeout rate (11.7%), and path length (9%) were less frequently reported, most likely because efficiency had less priority than feasibility in the studies considered.

Even though a wide range of social metrics exist, the most common among them were still applied more selectively. Separation distance (30%) and comfort or intrusion metrics (27%) were among the most frequently applied social metrics. Measures related to social naturalness (1.9%) or velocity smoothness (1.9%) were rare. Even rarer are metrics such as Zone-entering penalties, Group cohesion penalty, Group-aware rewards, Angle of approach, Elevator etiquette, Occlusion avoidance, Yielding/queuing, and Trajectory deviation, which appear only once or twice throughout the surveyed works.

The prioritization of technical feasibility over social awareness, combined with the absence of a unified set of metrics for social awareness, presents a significant gap. Additionally, the lack of standardized use of comfort, naturalness, and intrusion metrics makes direct comparison between algorithms difficult. It is therefore imperative to develop a common benchmark framework that balances technical performance with social acceptability.

### 4.2 Datasets

The importance of Datasets in SAN cannot be overstated. Datasets help train navigation algorithms to better and faster understand human behaviour, social interactions, and movement patterns. Some important datasets in this area include (SCAND) [144], a large-scale, first-person-view dataset. It contains 8.7 hours, 138 trajectories, and 25 miles of socially compliant, human tele-operated driving demonstrations. The input data includes data streams from 3D lidar, joystick commands, odometry, and visual and inertial information collected on two mobile robots in indoor and outdoor environments. It is well suited for imitation learning and sim-to-real transfer and allows policies to inherit socially compliant behaviors from real human teleoperation. The ETH [145] and UCY [146] datasets are commonly used in SAN for trajectory prediction. They include real-world pedestrian trajectories in public scenes and can also be used for motion prediction. These datasets are used in training trajectory prediction models and also provide data for setting up imitation learning when training a DRL navigation policy.

The THOR dataset [147] compiles human trajectory and eye gaze data using a motion capture system and 3D LiDAR. However, it has not been applied to DRL algorithms in Social navigation. THOR and SCAND [148] record the data of the robot navigating in the presence of humans. The SCAND dataset has the advantage of incorporating indoor and outdoor data and demonstrations, in contrast to THOR, which only has data from a controlled indoor environment. The datasets SocNav1 [149] and SocNav2 [150] use graph neural networks to learn social conventions using human feedback in simulation environments. They support evaluating human comfort and social acceptability.

Using datasets for training and testing socially aware DRL algorithms is advantageous, as it provides realistic scenarios for modeling human-robot interactions and speeds up the learning process. However, over-reliance on recorded datasets can also lead to problems, as the goal is to provide robust navigation algorithms that can perform effectively in diverse environments and learn a navigation policy even when encountering unfamiliar data and environments. Unfortunately, no existing dataset fully supports end-to-end DRL policy learning in social environments. Real-world datasets lack scale and diversity of social interactions, and synthetic datasets struggle with environmental validity. Future progress will thus be towards creating hybrid approaches that leverage simulation for scalable training with real-world demonstrations to embed social awareness accurately.

Table 8. Comparison of Datasets

| **Dataset** | **SCAND** | **ETH/UCY** | **THOR** | **SocNav1** | **SocNav2** |
|---|---|---|---|---|---|
| **Type** | Real-world (demonstration) | Real-world (trajectory data) | Real-world | Synthetic / real-world hybrid | Synthetic / real-world hybrid |
| **Purpose** | Imitation learning of socially compliant navigation policies | Modeling and learning human motion behavior and trajectory prediction | Human motion with robot present. Includes robot trajectory and gaze | Learning human discomfort evaluation of robot positions | Dynamic scenarios with goal-directed evaluation |
| **Size** | Moderate (8.7h, 138 trajectories, 25 miles | Small-medium (5 scenes of about 1–2 days each) | Moderate (rich annotations, unspecified duration) | Small (~9,280 samples) | Medium (53,600 dynamic social scenes) |
| **Sensing Modalities** | 3D LiDAR, RGB video (multi cameras), IMU, joystick commands, odometry | Video annotations. Stationary RGB camera | Robot and human trajectories, gaze, 3D LiDAR, obstacle maps | Static spatial layout: humans, objects, robot; discomfort score | Similar to SocNav1 but dynamic, adds navigation intent |
| **Application Scenarios** | Goal-specific indoor social navigation | Outdoor pedestrian navigation | Indoor environment | Evaluating robot disturbance | Evaluating robot trajectories |
| **Social Interaction Richness** | High (real human-robot interactions in crowds) | Small to Medium (dense, varied pedestrian trajectories) | Medium to High (human-robot in shared spaces) | Medium (static scenarios rating discomfort) | Medium (improves SocNav1 with dynamics and goal score) |
| **Strengths** | Authentic human-like navigation styles, sensor-rich, multi-robot, and applicable to IL | High realism; widely used for trajectory models | Precise annotations. Includes human gaze and social grouping | Structured. Labels human comfort directly | Adds dynamic interaction and goal evaluation |
| **Weaknesses** | Large data size (400 GB) limits accessibility; no DRL action-reward logs. | Lack of robot-specific interaction, limited volume and modalities | Limited scenario variety, not meant for policy training | Static and simplified with no dynamics or continuous trajectories | Synthetic, limited realism, and no trajectory-level data |

### 4.3 Learning Environments and Simulators

Testing SAN algorithms before deploying them in real-world mobile robots is highly recommended. Simulators allow rapid testing, training, and evaluation of algorithms before deployment. This allows for algorithm adjustments without the risks associated with testing in a real environment. The simulators incorporate human agents and accurately model their movements. The simulators include both two and three-dimensional frameworks. Among them is PedSimROS, a 2D crowd simulator based on the Social Force Model (SFM) and adapted for the ROS platform. As one of the earliest simulators for applying socially aware navigation in ROS, it is very popular. However, it is limited by being dependent on ROS1. Open AI Gym [151], as used in CrowdNav [23], is the most popular among the simulators and is widely used for DRL-based SAN methods. Its main limitation is in its lack of behavior realism due to using models rather than data-driven human motion.

Notably, SocNavGym [150] is a simulation environment specifically designed for DRL-based social navigation, capable of generating numerous navigation scenarios for training and testing DRL algorithms. Additionally, it is reconfigurable to different handcrafted and data-driven social reward signals, providing metrics to evaluate the robot's navigation performance. However, its abstracted nature means it doesn't simulate physics or deep sensor realism. Another simulator is the interactive Gibson (iGibson) simulator [152]. It runs on top of the PyBullet physics engine simulator and is used for 3D simulations and rendering of mobile robots interacting with other agents and their environments. It is not commonly used in the social navigation DRL domain, but its environment fidelity makes it powerful for future research. It was used for DRL-based socially aware navigation in [153]. Figure 6 shows 2D and 3D simulation environments.

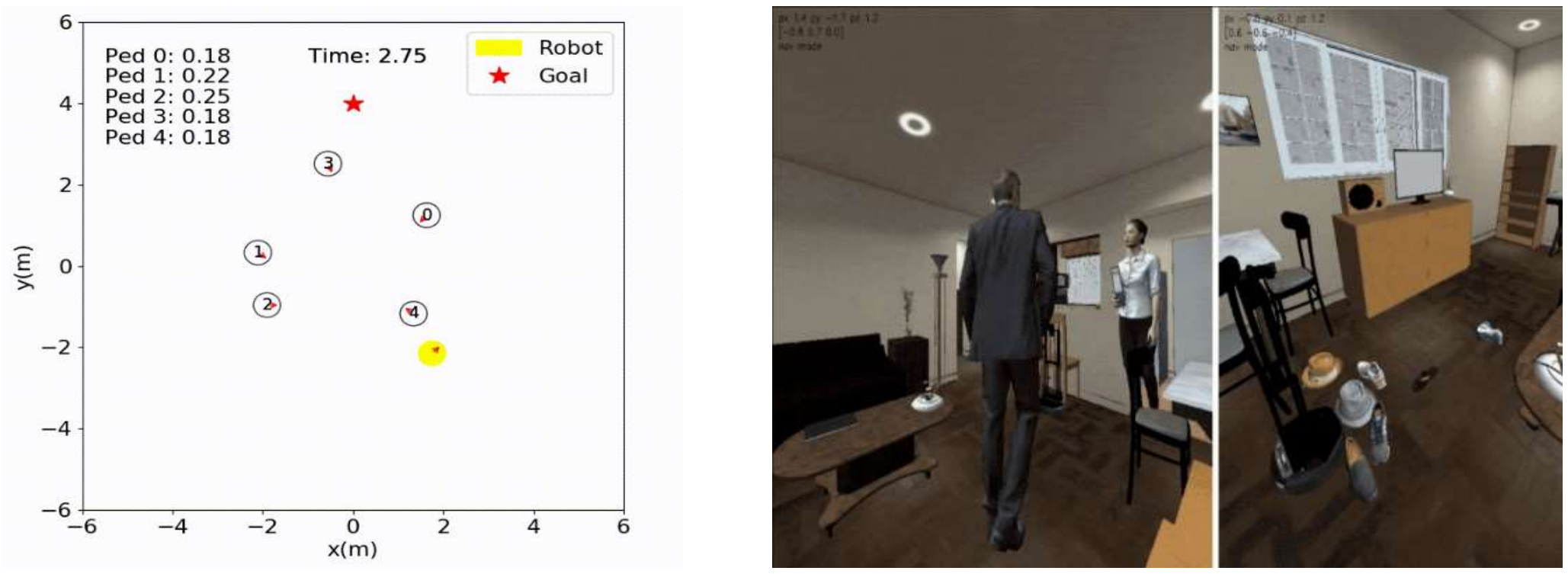

(a) (b)

**Fig. 6** Simulation Environments. **a** Open AI Gym [151]. **b** iGibson [152]

**Table 9 Comparison of Learning Environments**

| Simulator | PedSimROS | CrowdNav (OpenAI Gym) | SocNavGym | iGibson | Gazebo-ROS |
|---|---|---|---|---|---|
| Social Interaction Richness | Medium | Medium | Medium–High | High | Low–Medium |
| Human Motion Model | Social Force Model | ORCA or SFM (configurable) | Hand-crafted or data-driven | Physics-based mannequins and replayed trajectories | configurable |
| Sensor Simulation | Simulated point clouds | Relative position | Abstract state observations (configurable) | Full RGB-D, LiDAR | LiDAR, cameras, IMU |
| Integration with Real-World Data | Configurable | Not inherently integrated | Yes | Use real-world reconstructions/trajectories. | Can replay trajectories; integrate datasets via plugins. |
| Dimensionality | 2D | 2D | 2D | 3D | 3D |
| Performance Metrics | Customizable, includes: collisions, path efficiency, goal reaching rate | Success rate, collision rate, Navigation time, discomfort | Rich variety of metrics. Include: success and failure flags, efficiency, comfort/safety, kinematics | Interactive Navigation Score, Simulation performance metrics, and Task-based metrics | No built-in navigation/social metrics. Customizable, and defined by researchers. |
| Strengths | Real-time crowd behavior in ROS, group walking, sensor outputs compatible with robotics stacks | Standardized benchmark, reproducible, and integrates with DRL pipelines | Lightweight, fast, easy to extend, supports social reward tuning and scenario variety. | Realistic visual and sensory fidelity, supports sim-to-real transfer | Flexible, widely used, full-stack robotic integration, and extensible with human models |
| Limitations | Passive environment, deprecated ROS 1 integration, not gym-compatible, limited | Synthetic agent behaviors, lacks multi-modal sensors or deep social cues | Abstract environment, lacking complete physics or sensor realism | Computationally heavy, custom scenario setup required, less common in DRL social navigation papers | Heavy resource requirements, requires the user to integrate crowd models, and setup |
| | action-state feedback | | | | complexity can be high. |

Table 8 shows that SocNavGym is best used if a comprehensive evaluation of the algorithm is needed. CrowdNav is also strong for proxemics and crowd interaction evaluation but has narrower metrics. IGibson on the other hand is weaker on social metrics. The analysis of the table further reveals an imbalance in the use of 2D and 3D simulators. 2D simulators enable reproducibility and rapid DRL training due to their simplicity, while 3D simulators prioritize realism at the cost of scalability. This trade-off is vital in developing DRL-based SAN algorithms. Adoption of more recently developed simulators such as SocNavGym and iGibson which support data-driven human motion and richer social cues, have not yet caught up in the SAN research community compared to synthetic-model-based frameworks. Until this happens transferability of developed algorithms to real-world settings will remain a bottleneck. Future work will benefit from hybrid environments that combine DRL-friendly design with higher social realism, bridging the gap between efficient learning and human-centred evaluation.

### 4.4 Benchmarks

To correctly and fairly evaluate DRL-based socially aware methods, we need accurate benchmarks to serve as points of reference or baselines for comparison. To this end, researchers have proposed essential benchmarks, including SocialGym [142], a simulation-based approach well-suited for social navigation with RL approaches, and SocNavBench [143], a simulator with photorealistic capabilities based on real-world pedestrian data. Another important benchmarking tool is the PIC4rl-gym [154], which, although it is a non-socially aware DRL-based framework, combines ROS2 and Gazebo for navigating mobile robots in both indoor and outdoor scenarios, among other tasks.

SEAN 2.0 [155] was developed as an open-source system for training and benchmarking navigation policies in various social situations. It comprises robot tasks and environmental factors and uses logic-based classifiers. It has not been applied to benchmark DRL-based socially aware methods, but it can easily be adapted. SRPB [156] is another benchmark for quantitative evaluation. It incorporates both traditional and human-aware trajectory planners, extending to real-world environments beyond ROS simulations. This benchmark can also be used to test DRL-based socially aware navigation algorithms.

### 4.5 Simulations vs Real-world implementations

Using simulations in DRL-based SAN allows for safe and efficient training and testing of algorithms in the navigation environment. Because the environment is controlled, the simulations are easily repeatable, and the algorithms can be fine-tuned under the same conditions without the risk of damage to hardware or human accidents. However, these simulations must be made to accurately model the real world and the interactions that happen within it; otherwise, when the developed algorithms are transferred to physical robots for deployment in environments where robots are required to navigate socially, they may not function accurately. An example of a DRL-based SAN framework [86] that is simulated and then subsequently deployed in the real world is shown in Figure 7.

Several challenges occur when transferring from simulation to hardware. Among them is the diverse nature of human behaviours, in which some works have implemented deviations in pedestrian models and value networks that are continuously updated to enhance the learning process to estimate this pedestrian unpredictability. These changes aid in lowering collision rates and enhancing navigation safety in strange environments [141]. Memory-enhanced frameworks have also been developed to solve the problem of a lack of real-time communication and interaction with pedestrians in simulations. They store important information about the environment and model sequential dependencies, transferring the frameworks efficiently to real pedestrian scenarios [79]. Because socially aware robot navigation methods with DRL generally have high-dimensional state and action spaces, and most mobile robotic platforms have limited onboard computational power, some works tend to simplify it to a small set of actions. This is disadvantageous, and other approaches have aimed to tackle this by using methods like human guidance [157], Object Q-learning [158], and action elimination [159].

Another challenge is the design of reward functions, which may be unspecified, multi-objective, risk-sensitive, or even different from those used in real-world training and testing. Some works utilize simplified real-world reward functions that are limited to one parameter, such as user feedback or predefined numeric values. In contrast, the simulated world reward functions are more complex, including several parameters. The reward functions tend to be complex and require intricate human-crafted designs. Transformable Gaussian reward functions have been proposed to reduce hyperparameter tuning and improve learning rates, particularly in crowded environments [36]. Only about half of the surveyed works have implemented their algorithms physically on a mobile robotic platform to verify their effectiveness. The TurtleBot is the most common robotic platform used due to its ease of integration with various algorithms, open-source nature, and availability of numerous resources; some researchers have also utilized the Fetch robot and the Jackal robot. However, these successes often rely on simplified crowd models and hand-tuned domain randomization. In contrast, real-world failures are linked to unmodeled human behaviors, perception noise, and occlusion effects that violate the simulation's assumptions. In Figure 7, the rendered simulation output of [86], which is an occlusion avoidance method is shown together with a snapshot of their video demonstration of the implemented framework after being transferred to a Turtlebot robot.

Bridging the sim-to-real gap in DRL-based social navigation is a promising and emerging field, and several approaches have been adopted by researchers to solve this issue. An approach of using Hybrid learning along with an optimizer was implemented in [114] to enforce kinematic/safety constraints when the learning frontend is uncertain. It combined learnt decisions with an optimization/safety layer to preserve sim performance in reality. Likewise a Low-dimensional, dynamics-aware observation framework, [102] was developed to build the policy around physically feasible control primitives to make transfer easier among other methods. Datasets such as ETH, UCY, and SCAND also provide a foundation for learning human motion behaviors that improve real-world transferability.

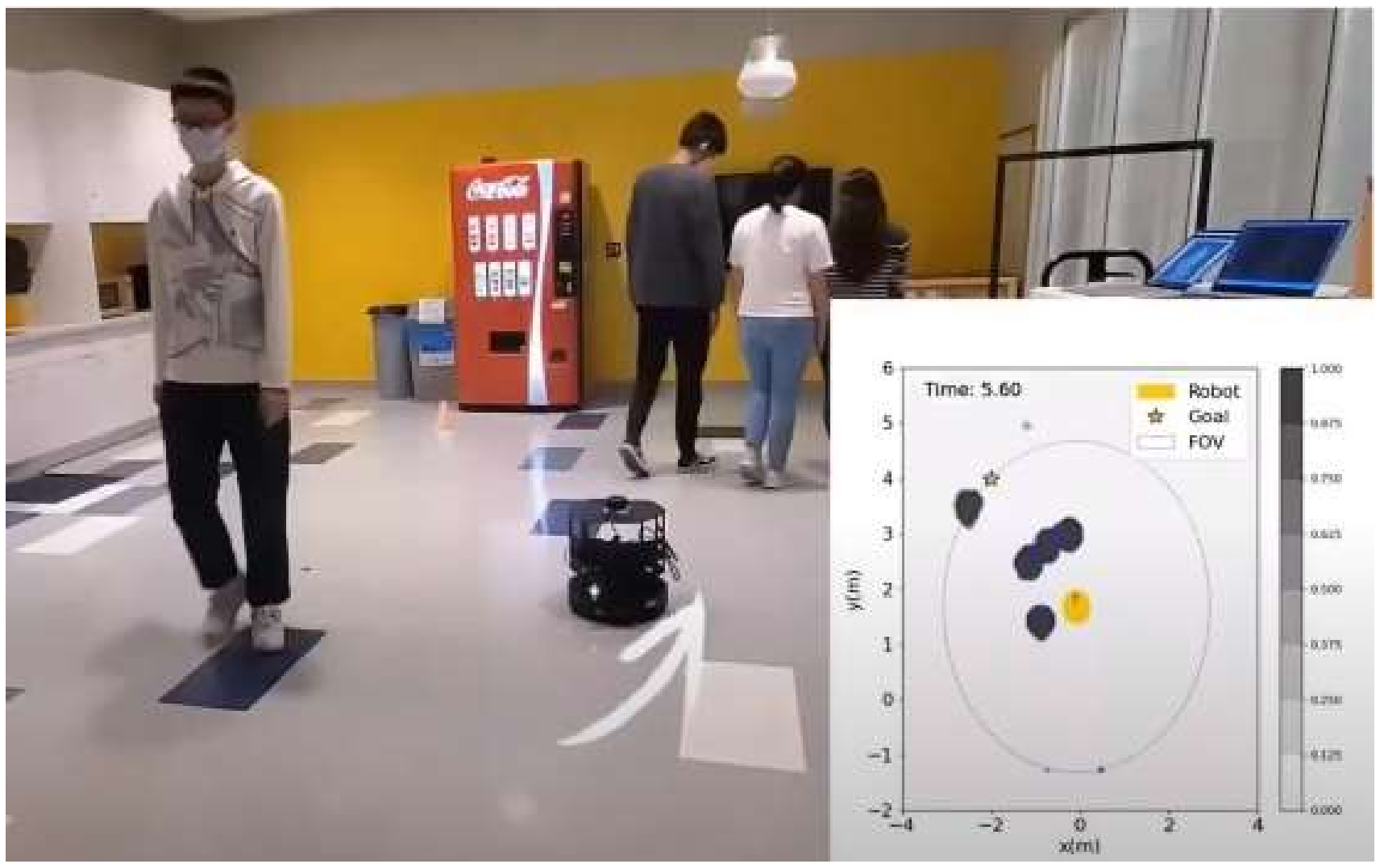

**Fig. 7** An illustration of an implementation of a simulated framework in a real-world scenario [86]

### 4.6 Standardization of Evaluation Protocols

Socially aware navigation is a relatively new and rapidly evolving field. Despite recent progress, evaluation remains a challenge. Current evaluation mechanisms are unstandardized and fragmented, with no unified metrics that measure both technical performance and social compliance, as seen from the various types of metrics in Section 4.1. To move toward standardization, a clear roadmap for developing and validating social metrics is needed. We suggest identifying a small, core set of reproducible quantitative metrics, such as personal space compliance (PSC), time-to-collision (TTC), clearance distance, and velocity smoothness, to complement traditional measures like success rate and path efficiency. These baseline metrics can initially be validated in simulation using diverse crowd scenarios and pedestrian behavior models to ensure consistency and reproducibility.

Building upon this, we suggest human-centered validation to link quantitative metrics with subjective human feedback through controlled studies or annotated datasets. Combining data that includes objective technical metrics with corresponding human ratings on comfort, safety, and legibility can help researchers to train and standardize learned social metrics that predict perceived social acceptability. The learned model obtained from this can then be benchmarked for correlation and generalization across different environments and datasets, thus producing social metrics that are applicable across virtually any simulator. Conclusively, we recommend adopting scenario cards and a shared metric schema across research groups to help establish a standardized evaluation protocol. This will undoubtedly facilitate direct comparison and cumulative progress in socially aware navigation research.

## 5. CHALLENGES AND FUTURE DIRECTIONS

This section explores the challenges of implementing DRL in Socially Aware navigation and proposes future research directions.

### 5.1 CHALLENGES

The challenges of integrating DRL to enhance socially aware navigation for mobile robots are discussed in this section.

#### 5.1.1 Uncertainty and Unpredictability

The navigation of robots running DRL socially aware algorithms can be affected by uncertainty and unpredictability. The robots may struggle to predict and react to sudden changes in human movement patterns. Additionally, the robots can struggle with delays because of computational delays, given that simulations are instantaneous. At the same time, in the real world, the robot might experience a lag or slow reaction time, possibly due to delays in the actuators and sensors.

#### 5.1.2 Real-World Applicability

One of the biggest challenges in running DRL-based SAN algorithms on physical robots is Sim-to-real transfer. This is the ability to transfer learned models and algorithms to physical robots for deployment in the real world. Many works have developed DRL socially aware algorithms only in simulation and have not deployed them to physical environments. This dramatically limits the certainty with which we can accept that these works accurately address the primary issue of having robots navigate in human environments without posing a danger or discomfort to humans. Thus, more real-world implementations on physical robots would inform whether the algorithms are computationally compatible with the type of hardware currently available for mobile robots or if the algorithms are too demanding for robots. For works that implement the algorithm in the real world, they struggle severely to replicate the same results as in simulations because simulations cannot fully and accurately capture all the intricacies of the physical environment.

#### 5.1.3 Computational Complexity

DRL algorithms are usually very computationally demanding. To achieve socially aware navigation, they must handle high-dimensional continuous state and action spaces, which provide more accurate representations of how humans and robots interact in the real world, compared to discrete state and action spaces. These often require complex neural networks, which are computationally expensive to train and deploy. At the same time, most robots have only small onboard computers, which have very limited computational power.

#### 5.1.4 Generalization

Generalization refers to the ability of algorithms to perform well in environments or scenarios they haven't been explicitly trained on. Robots should be able to use the learned policies in new environments. However, the algorithms tend to perform well in a specific type of crowd or simulation but fail when faced with unfamiliar patterns or dynamics. This is also the case when they come across variability in human behavior and interaction styles across different cultural or environmental contexts.

#### 5.1.5 Scalability

One of the challenges in deploying DRL in robots for SAN is scalability, as most of the papers reviewed implemented multi-robot DRL on only a few robots and tens of robots. This makes it challenging to implement in the real world. One reason for this is that such algorithms require large amounts of data and correspondingly large amounts of computational resources. This problem becomes magnified, especially when it is desired that all the robots work together cooperatively. Therefore, there is a need to develop lightweight techniques and integrate them with the latest techniques in accelerated computing to fully integrate multiple robots into human spaces.

#### 5.1.6 Reproducibility and Need for Open-Source Benchmarks

Another barrier to progress is reproducibility; very few studies and papers have provided enough resources and details to make their work reproducible and build upon. Similarly, most studies currently rely on heterogeneous datasets, simulation frameworks, and metrics, making direct comparison difficult. Open benchmarks with common datasets, evaluation protocols, and scenario descriptions would enable fair comparison and validation thus improving reproducibility and transparent experimentation.

### 5.2 FUTURE DIRECTIONS

Building upon the challenges of socially aware navigation with DRL, the following research directions are highlighted below to advance this area of study.

1. In the future, increasing the efficiency of the DRL algorithms is a promising direction to make them more computationally efficient. Finding ways to enhance the convergence rate of the RL algorithms to enable real-time decision-making is a fundamental requirement
2. Future work should focus on generalization to enhance performance of algorithms across various environments and application scenarios.
3. Improving the scalability of the DRL-based SAN algorithms to be efficiently deployed in more diverse and complex scenarios and with multiple robots.
4. More distinct and accurate representation of agents and their behaviours, without the assumption that all agent behaviours are homogeneous is needed. Robust and adaptive behaviour models should be developed to account for factors that may affect how robots navigate around humans, such as age, personality, gender, cultural differences, and prior experience with robots [160].
5. Future research can explore uncertainty-aware frameworks to accurately infer context-specific reward functions by leveraging techniques such as inverse reinforcement learning
6. Making transferable policies across different robot types to aid rapid deployment and applying these algorithms to a more diverse range of robots, e.g., legged, aerial, or mobile robots with different kinematics and dynamics.
7. More research is needed to explore the cooperative navigation of socially aware mobile robots using deep reinforcement learning, where the robot and humans cooperate in their navigation goals explicitly and implicitly, and specifically for deployment with multiple robots and in diverse human environments.

8. Creating more robust simulation environments that incorporate human diversity, unmodelled human behaviour, and uncertainty, which will be more accurate in capturing the real-world environment. This ensures minimal performance gaps when the algorithms are deployed on physical mobile robots.
9. Future SAN systems should focus on developing hybrid learning strategies that are also interpretable. This dual approach will deliver both reliable performance and understandable, predictable policies.
10. Embedding ethical and HRI principles into learning objectives, such as fairness, politeness, and transparency, will be essential for socially acceptable deployment in future real-world contexts.
11. An interesting research direction will be the study of how DRL-based SAN systems may begin to shape human social behavior as they adapt to it.

## 6. CONCLUSION

This survey paper has provided an overview of the advancements, techniques, and challenges in applying deep reinforcement learning to socially aware navigation of mobile robots. The applications of DRL to SAN are a promising and continuously expanding field. This survey highlights the use and applicability of several DRL types, such as value-based, policy-based, and actor-critic-based algorithms, for specific considerations. Additionally, the role of different types of neural networks in relationship and interaction representation is thoroughly highlighted. Several challenges, such as computational complexity, generalization across new environments, and the 'sim-to-real' gap, are also highlighted. In the future, researchers could explore improving generalization, optimizing algorithms to make them more computationally efficient, exploring implementation on other robot types, and considering more diverse human behaviour. We believe this work will serve as a good foundation and a stepping stone to understanding the role of DRL in the socially aware navigation of mobile robots.